\pdfoutput=1
\documentclass[11pt]{article}
\PassOptionsToPackage{hyphens}{url}
\usepackage[table]{xcolor}
\usepackage{techreport}
\def\TechReport{}

\usepackage{amsfonts}
\usepackage{inconsolata}
\usepackage{url}
\usepackage{booktabs}
\usepackage{amsmath}
\usepackage{amssymb}
\usepackage{wrapfig}
\usepackage{graphicx}
\graphicspath{{images/}}
\usepackage{subcaption}
\usepackage{float}
\usepackage{adjustbox}
\usepackage{array}
\usepackage{listings}
\usepackage{siunitx}

\definecolor{contextblue}{RGB}{230,240,255}
\definecolor{skillgreen}{RGB}{230,248,235}
\definecolor{skillpurple}{RGB}{239, 204, 236}

\lstdefinestyle{skillstyle}{
    basicstyle=\ttfamily\small,
    breaklines=true,
    frame=single,
    backgroundcolor=\color{gray!5},
    columns=fullflexible,
    keepspaces=true
}
\definecolor{contextbar}{HTML}{EE5960}
\definecolor{contextfill}{HTML}{FFE1E3}
\definecolor{questionbar}{HTML}{5C95DB}
\definecolor{questionfill}{HTML}{EAF2FE}

\newtcolorbox{infobox}[3][]{%
  enhanced,
  width=\linewidth,
  colback=#3,
  colframe=#2,
  colbacktitle=#2,
  coltitle=white,
  fonttitle=\sffamily\bfseries\large,
  title={#1},
  boxrule=0.7pt,
  arc=2mm,
  left=4mm,
  right=4mm,
  top=3mm,
  bottom=3mm,
  boxsep=0pt,
  before skip=0pt,
  after skip=3mm
}

\newcommand{\best}[1]{\textbf{#1}}
\newcommand{\second}[1]{\underline{#1}}
\newcommand{\equalcontrib}{\textsuperscript{*}}
\newcommand{\projectlead}{\textsuperscript{\ensuremath{\dagger}}}
\newcommand{\corrauthor}{\textsuperscript{\ensuremath{\ddagger}}}

\newcommand{\floatsNormal}{}

\title{Parametric Skills}
\reportorg{Shanghai AI Laboratory}

\authors{%
Xuan Zhao\textsuperscript{1,2}\equalcontrib,\quad
Haonan He\textsuperscript{1,2}\equalcontrib\projectlead,\quad
Qingyu Yang\textsuperscript{1,3}\equalcontrib,\quad
Minglei Li\textsuperscript{1,4},\\[2pt]
Zelin Tan\textsuperscript{1,2},\quad
Bo Wan\textsuperscript{2},\quad
Jingqi Ye\textsuperscript{1,2},\quad
Chen Zhang\textsuperscript{1},\quad
Shuyue Hu\textsuperscript{1},\quad
Peng Ye\textsuperscript{1,4,5}\corrauthor}

\affiliations{%
\textsuperscript{1}Shanghai Artificial Intelligence Laboratory\quad
\textsuperscript{2}University of Science and Technology of China\\
\textsuperscript{3}KTH Royal Institute of Technology\quad
\textsuperscript{4}Fudan University\quad
\textsuperscript{5}The Chinese University of Hong Kong}

\authornotes{\equalcontrib Equal contribution\quad
\projectlead Project lead\quad
\corrauthor Corresponding author}

\metaline{%
  \metaitem{\faEnvelope}{Correspondence}{\href{mailto:yepeng@pjlab.org.cn}{\texttt{yepeng@pjlab.org.cn}}}\qquad
  \metaitem{\faCalendar*}{Date}{\today}}
\metaline{%
  \metalink{\faGlobe}{https://hhnqqq.github.io/ParametricSkills/}{Project Page}\qquad
  \metalink{\hficon}{https://huggingface.co/DiaryOfBacon/ParametricSkills}{Hugging Face}}

\begin{document}

\begin{abstract}
For LLMs, skills
are textual recipes encoding mature problem-solving experience. Despite widespread deployment, their utility is limited by the model's ability to comprehend and follow skill instructions. To address this limitation, we propose \textsc{ParametricSkills}, a framework that parameterizes textual skills at test time, enabling context-free management of the full skill lifecycle. Specifically, we first construct a large-scale and high-quality skill collection, and then synthesize single-turn and multi-turn skill-exploitation trajectories. Using this data, we then train a hypernetwork that parameterizes both the content and the exploitation methodologies of skills by converting textual skills into LoRA adapters. To facilitate the real-world application of our framework, we further align the hypernetwork on complex, long-horizon trajectories collected with OpenCode, enabling the models to operate reliably in realistic interactive workflows with \textsc{ParametricSkills}. Across evaluations, \textsc{ParametricSkills} effectively parameterizes skills, yielding substantial gains in task completion. More notably, as parametric skills accumulate as adapters rather than crowding the context, they open a preliminary but promising path toward continual learning at test time.
\end{abstract}
\maketitle

\noindent\begin{minipage}{\textwidth}
  \centering\captionsetup{skip=3pt,hypcap=false}
  \includegraphics[width=0.9\textwidth]{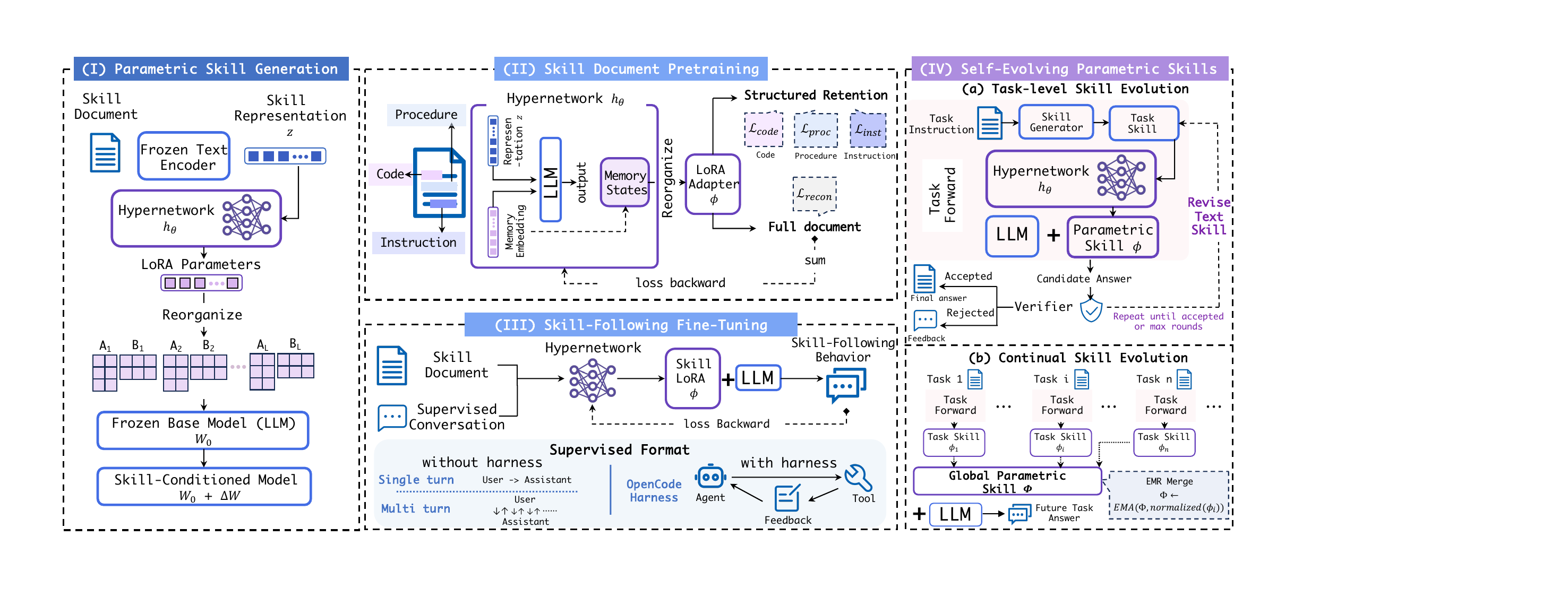}
  \captionof{figure}{Overview of \textsc{ParametricSkills}.
    (I) Parametric skill generation;
    (II) skill-reconstruction pretraining with structured reweighting;
    (III) skill-exploitation fine-tuning on single- and multi-turn examples
    without a coding-agent harness (Stage~2) and OpenCode trajectories
    with a harness (Stage~3);
    (IV) task-level skill evolution and continual learning.}
  \label{fig:overall_pipeline}
\end{minipage}
\trcontents

\begingroup
\RenewEnviron{figure}[1][]{}
\section{Introduction}
Agentic Large Language Models (LLMs)~\citep{zeng2026glm,deepseekai2026deepseekv4} have demonstrated remarkable capabilities in complex real-world tasks such as software engineering~\citep{yang2024swe, wang2025openhands}, scientific research~\citep{phan2025humanity, wang2026frontierscience,yang2026aris}, and Olympiad-level reasoning~\citep{yu2025hipho, huang2025winning, li2026achieving}. A key enabler of the capabilities of LLMs-driven agentic systems is skills~\citep{anthropic2025equipping, zhou2026comprehensive}, structured textual recipes that encapsulate mature problem-solving experience and other procedural knowledge, either manually authored by experts or extracted from previous task trajectories. By maintaining a skill collection and retrieving from it, agents can add, evolve, and reuse proven strategies across tasks rather than reasoning from scratch, thereby achieving better and more stable task completion quality.

\begin{figure}[t]
    \centering \includegraphics[width=\textwidth]{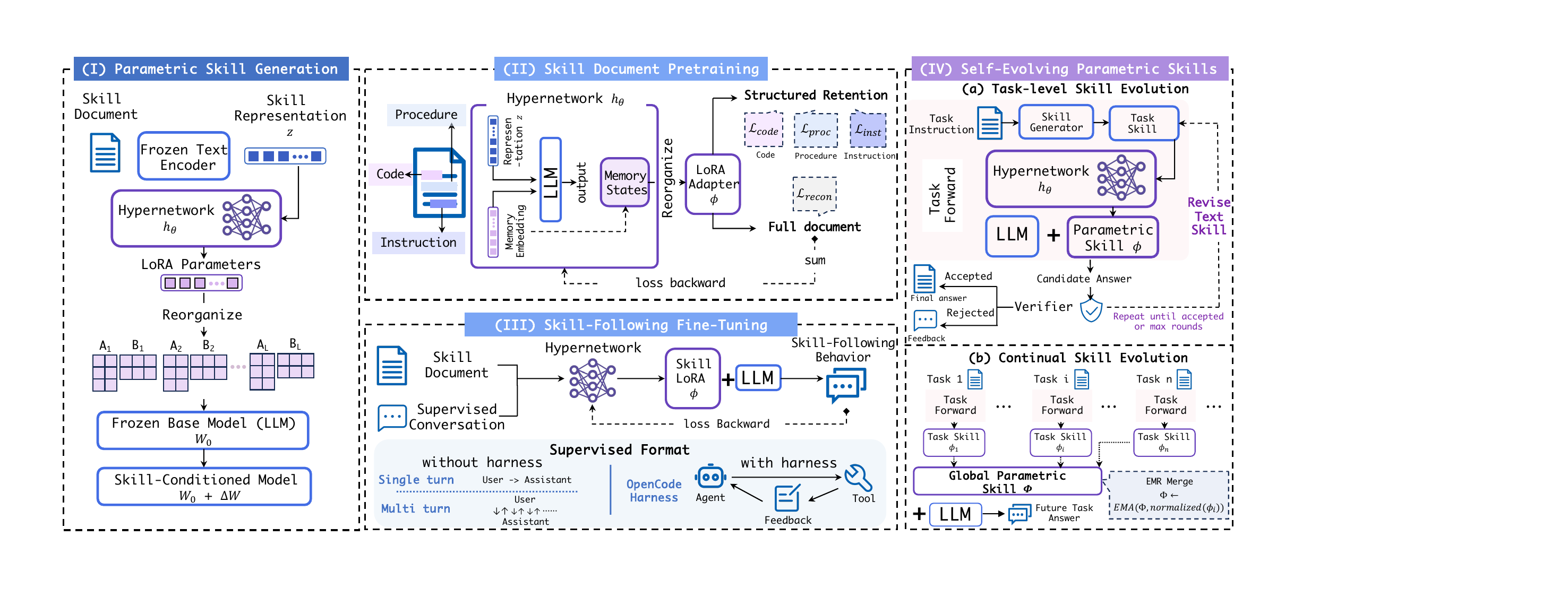}
    \caption{Overview of \textsc{ParametricSkills}.
    (I) Parametric skill generation;
    (II) skill-reconstruction pretraining with structured reweighting;
    (III) skill-exploitation fine-tuning on single- and multi-turn examples
    without a coding-agent harness (Stage~2) and OpenCode trajectories
    with a harness (Stage~3);
    (IV) task-level skill evolution and continual learning.}
    \label{fig:overall_pipeline}
\end{figure}

However, the prevailing paradigm of in-context skill suffers from two fundamental limitations. First, the effectiveness of textual skills is bottlenecked by the model's ability to comprehend and follow complex instructions. Open-source models~\citep{grattafiori2024llama, yang2025qwen3} of varying scales, despite being widely deployed, largely fail to fully exploit skills through in-context learning (ICL), especially in long-context, complex task-completion scenarios where vital instructions are hard to locate and follow~\citep{liu2024lost,liu2025comprehensive}. Moreover, since prior work has shown that ICL underperforms the post-training methods, 
and this gap widens sharply as model size decreases~\citep{lu2026skill0, shi2026skill1, xia2026skillrl, wang2026skill-sd},
the lack of sufficient capability in understanding and following skills may lead to suboptimal task outcomes. Second, optimizing or evolving skills in text space is inherently decoupled from the learning of the models themselves. Recent approaches to skill evolution, such as EvoSkill~\citep{alzubi2026evoskill}, CoEvoSkills~\citep{zhang2026coevoskills}, and SkillOpt~\citep{yang2026skillopt}, iteratively rewrite natural-language skills based on failure analysis, co-evolving verifiers, and other mechanisms, enabling enhanced downstream performance without updating the models. However, these text-based methods remain detached from the intrinsic capabilities of LLMs, as the models themselves remain unoptimized. 

In light of these observations, we turn our attention to representing skills in parameter space, as parametric skills eliminate the model's dependence on locating and following intricate textual instructions, and parametric skill evolution becomes inherently coupled with model learning.
Therefore, we draw on recent progress in test-time parameter generation represented by hypernetwork-driven parameter generation~\citep{ha2016hypernetworks,chauhan2024brief}. For LLMs, most studies use hypernetworks to generate LoRA~\citep{hu2021lora} adapters. For example, Text-to-LoRA~\citep{charakorn2025text} and LoRAGen~\citep{xiao2025lora-gen} use a hypernetwork to generate task-specific adapters to mitigate the training cost of fine-tuning LLMs on downstream tasks; Doc-to-LoRA~\citep{charakorn2026doc} and SHINE~\citep{liu2026shine} leverage hypernetworks to convert context into LoRA adapters, thereby eliminating the need for storing the KV cache of context. Inspired by these advances, we propose applying test-time parameter generation to the full lifecycle of agentic skills, encompassing creation, composition, deployment, exploitation, and continuous evolution, thereby directly addressing both the skill instruction-following bottleneck and the decoupling of skill evolution from model learning.

We present \textsc{ParametricSkills}, a framework that converts textual skills into parametric adapters 
through a hypernetwork (Figure~\ref{fig:overall_pipeline}(I)). To train the hypernetwork to faithfully parameterize both the content of skills and the methodologies for exploiting them, we construct a skill collection covering 12 domains (e.g., software and AI \& LLM), with skills either downloaded online or summarized from real-world trajectories. We also synthesize and collect skill-exploitation trajectories with and without a coding-agent harness across various skill-use scenarios. 
We then pretrain the hypernetwork on the skill collection and fine-tune it on the skill-exploitation trajectories (Figure~\ref{fig:overall_pipeline}(II\&III)). 
Evaluations on ALFWorld~\citep{shridhar2020alfworld} and SkillEval~\citep{tan2026skt} demonstrate the effectiveness of \textsc{ParametricSkills} on Qwen3-8B~\citep{yang2025qwen3}, Qwen3.5-4B, and Qwen3.5-9B~\citep{qwen3.5}. Specifically, on ALFWorld~\citep{shridhar2020alfworld}, \textsc{ParametricSkills} achieves average success rates of 93.43\%, 90.15\%, and 95.26\% on the three models, respectively, improving over ICL by 17.52, 65.70, and 31.76 points. On SkillEval~\citep{tan2026skt}, it achieves scores of 48.89, 57.83, and 63.12, surpassing ICL by 6.41, 0.04, and 2.16 points on the three models, respectively. Beyond benchmark performance, we demonstrate self-evolution of parametric skills through text-space revision, parametric compilation, and task-level verification, where only successfully verified candidates are accepted. We further explore continual learning by accumulating accepted parametric skills into a global parametric skill (Figure~\ref{fig:overall_pipeline}(IV)). In summary, our contributions are:
\begin{itemize}
\item We construct a large-scale skill collection containing 106.4k skills from two main sources: skills crawled from the web and skills summarized from real-world agentic trajectories. We then build a high-quality large-scale dataset of skill-use trajectories by applying skills from the collection.
\vspace{-2mm}
\item By training a hypernetwork on the aforementioned data, we propose \textsc{ParametricSkills}, a framework that internalizes both the content and exploitation methodologies of skills into parametric adapters at test time. Evaluations on ALFWorld and SkillEval demonstrate that
\textsc{ParametricSkills} improves over ICL on Qwen3-8B, Qwen3.5-4B, and Qwen3.5-9B substantially.
\vspace{-2mm}
\item We demonstrate that \textsc{ParametricSkills} can be seamlessly integrated with self-evolving pipelines of textual skills by evolving skills in text space. 
We further show that \textsc{ParametricSkills} can be a promising gateway for continual learning of LLMs by continually merging new parametric skills summarized from task experiences into the parameter spaces of LLMs.

\end{itemize}

\endgroup
\section{Preliminaries}
\subsection{LoRA}
LoRA enables efficient fine-tuning of models by injecting low-rank update matrices into linear layers. Let the base model be denoted by \(\mathcal{M}_{\Theta_0}\), whose full set of pretrained parameters is \(\Theta_0\). For each selected linear layer \(l \in \mathcal{L}\), LoRA freezes the pretrained weight \(W_0^{(l)} \in \mathbb{R}^{d_{\text{out}}^{(l)} \times d_{\text{in}}^{(l)}}\) and learns a low-rank update
\begin{equation}
  W'^{(l)} = W_0^{(l)} + \Delta W^{(l)}, \quad \Delta W^{(l)} = B^{(l)} A^{(l)},
\end{equation}
where \(A^{(l)} \in \mathbb{R}^{r^{(l)} \times d_{\text{in}}^{(l)}}\) and \(B^{(l)} \in \mathbb{R}^{d_{\text{out}}^{(l)} \times r^{(l)}}\) are low-rank factors with rank \(r^{(l)} \ll \min(d_{\text{in}}^{(l)}, d_{\text{out}}^{(l)})\). 
At inference, LoRA adapters are natively plug-and-play: they can be added, removed, or swapped without modifying the base model \(\mathcal{M}_{\Theta_0}\).

This plug-and-play property makes LoRA an ideal substrate for representing discrete agentic skills as continuous, composable parameters. Specifically, for a skill \(s\), we can convert it into parametric space by generating a LoRA collection
\(
  \Phi(s) = \{\phi^{(l)}(s)\}_{l \in \mathcal{L}},
\)
where each \(\phi^{(l)}(s)\) corresponds to the LoRA factors for layer \(l\), i.e., \(\phi^{(l)}(s) = (A^{(l)}(s), B^{(l)}(s))\). After loading the LoRA collection, we denote the parametric-skill-conditioned model by
\begin{equation}
  \mathcal{M}_{\Phi(s)}(\cdot)
  \triangleq
  \mathcal{M}_{\Theta_0}(\cdot; \Phi(s)).
\end{equation}

\subsection{Hypernetwork-driven text to LoRA}
A hypernetwork \(h_\theta\) is a neural network that generates the weights of another network. We adopt this paradigm as the core of \textsc{ParametricSkills} to parameterize skills. Let \(s\) denote a skill description in text. We embed \(s\) using a frozen text encoder \(g(\cdot)\) into a skill representation \(z = g(s) \in \mathbb{R}^{d_z}\). A hypernetwork \(h_\theta\), parameterized by \(\theta\), then maps \(z\) to a collection of LoRA parameters:
\begin{equation}
  \Phi(s) = h_\theta(z) = h_\theta\bigl(g(s)\bigr)
  = \{\phi^{(l)}(s)\}_{l \in \mathcal{L}},
\end{equation}
where \(\phi^{(l)}(s)\) is reshaped into \(A^{(l)}\) and \(B^{(l)}\) and loaded into the base model \(\mathcal{M}_{\Theta_0}\). This yields the parametric-skill-conditioned model \(\mathcal{M}_{\Phi(s)}(\cdot)\).
Once \(h_\theta\) is trained, generating a new parametric skill requires only a single forward pass of \(h_\theta\). 

\section{\textsc{ParametricSkills}}

\subsection{Training Pipeline of \textsc{ParametricSkills}}

After training on the general pretraining pipeline in SHINE~\citep{liu2026shine}, we train the hypernetwork of \textsc{ParametricSkills} in three stages. Stage 1 optimizes a textual reconstruction objective over skills, with the LLM conditioned on the generated parametric skills. Stages 2 and 3 share the goal of teaching the model to use parametric skills to solve tasks, using skill-exploitation trajectories without and with a coding-agent harness, respectively. Throughout all three stages, the backbone model remains frozen, while gradients propagate through the generated adapters to update the hypernetwork. Appendix~\ref{app:three-stage-examples} illustrates the data formats of all three stages with training examples.

\subsubsection{Stage 1: Skill-Reconstruction Pretraining}

This stage aims to enable the hypernetwork to faithfully convert the content of skills to support subsequent skill exploitation. Specifically, given a textual skill \(s\), the hypernetwork generates a corresponding LoRA collection \(\Phi(s)=h_\theta(g(s))\). The adapted language model learns to reconstruct the textual skill through teacher forcing. Formally, the adapted model, conditioned on the generated LoRA collection \(\Phi(s)\), the reconstruction prompt \(P_S\), and the first \(t-1\) tokens of the textual skill \(s_{<t}=(s_1,\dots,s_{t-1})\), predicts the \(t\)-th token:\(
  p_\theta(s_t \mid P_S, s_{<t})
  \triangleq
  \mathcal{M}_{\Phi(s)}
  \bigl(s_t \mid P_S, s_{<t}\bigr).
\)
Writing the token loss as \(\ell_t(s;\theta)=-\log p_\theta(s_t\mid P_S,s_{<t})\), the reconstruction loss is
\begin{equation}
  \mathcal{L}_{\text{recon}}(\theta)
  =
  \mathbb{E}_{s\sim\mathcal{S}}
  \frac{1}{|s|}
  \sum_{t=1}^{|s|}
  \ell_t(s;\theta).
\end{equation}

Beyond simple reconstruction, faithful internalization must preserve the critical details that govern how a skill should be applied. Such details may occupy only a small fraction of a document, limiting their contribution to the overall objective. We therefore introduce a structured reweighting objective that gives additional weight to these spans. For a textual skill \(s\), we identify three categories of key spans \(\{\mathcal{K}_c(s)\}_{c\in\mathcal{C}}\), where
\(
\mathcal{C}=\{\text{code},\text{procedure},\text{instruction}\},
\)
corresponding to code blocks, procedural steps, and behavioral instructions. The complete skill \(s\) conditions adapter generation, and per-token losses are computed once by the forward pass used in \(\mathcal{L}_{\text{recon}}\). The reweighting objective reuses the same token losses \(\ell_t(s;\theta)\). For each category \(c\), we average losses within each span and then across spans:
\begin{equation}
\mathcal{L}_{\text{reweight}}^{(c)}(s;\theta)
=
\frac{1}{|\mathcal{K}_c(s)|}
\sum_{k\in\mathcal{K}_c(s)}
\frac{1}{|k|}\sum_{t\in k}\ell_t(s;\theta).
\end{equation}
Here, \(k\) is the set of token positions in an annotated span. If category \(c\) is absent, its loss is the full-document mean \(\frac{1}{|s|}\sum_t\ell_t(s;\theta)\). We combine the category losses using normalized weights:
\begin{equation}
\mathcal{L}_{\text{reweight}}(\theta)
=
\mathbb{E}_{s\sim\mathcal{S}}
\sum_{c\in\mathcal{C}}
\lambda_c \mathcal{L}_{\text{reweight}}^{(c)}(s;\theta),
\end{equation}
where \(\lambda_c\ge 0\) and \(\sum_{c\in\mathcal{C}}\lambda_c=1\). We combine the reconstruction loss with the reweighting loss:
\[
\mathcal{L}_{\text{pretrain}}(\theta)
=
(1-\beta)\mathcal{L}_{\text{recon}}(\theta)
+
\beta \mathcal{L}_{\text{reweight}}(\theta),
\]
where \(\beta\in[0,1]\) is a hyperparameter. Appendix~\ref{app:reconstruction-ablation} details the ablation results.

\begin{figure}[t]
    \centering
    \includegraphics[width=\linewidth]{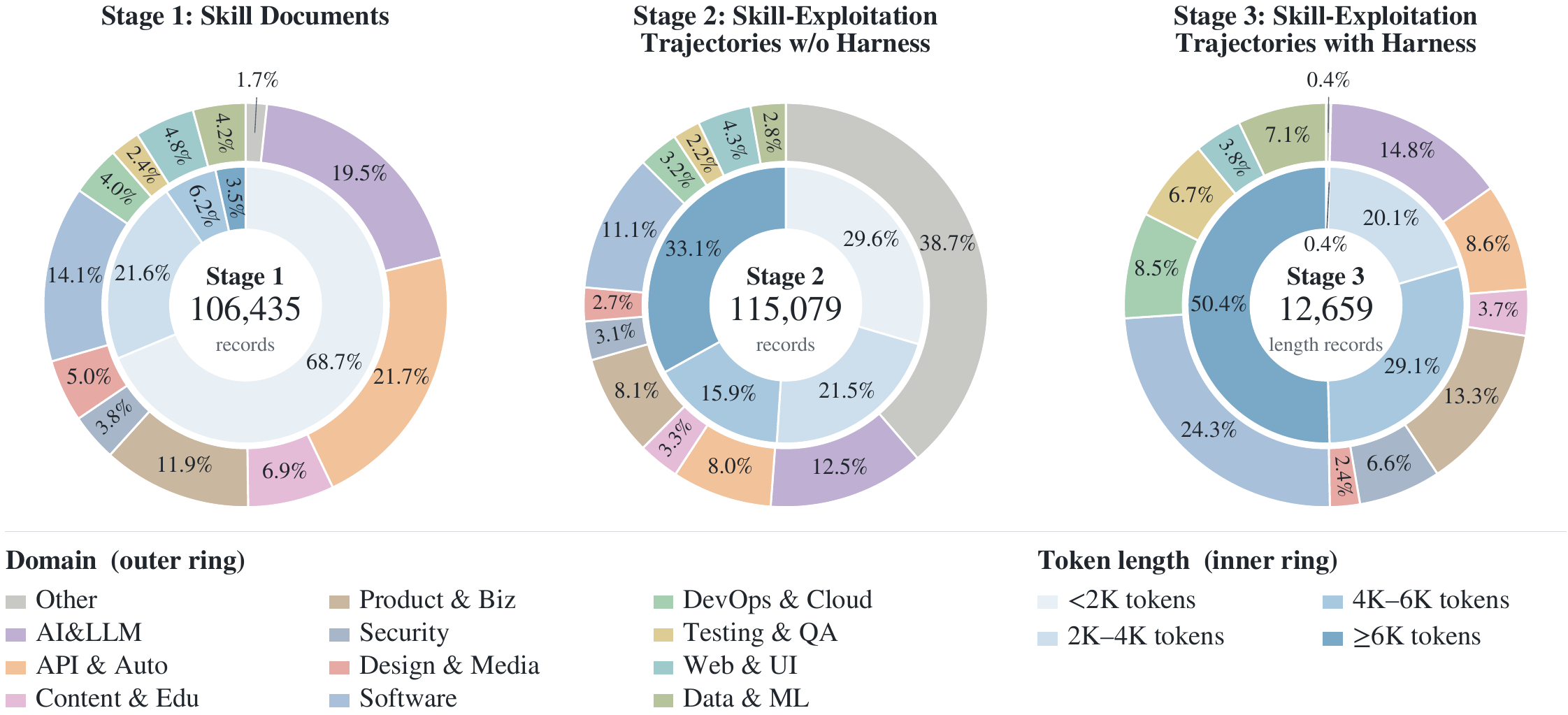}
    \caption{Domain and token-length composition across the three training stages.
    Outer rings show primary-use domains; inner rings show token-length distributions.}
    \label{fig:skills}
\end{figure}

\subsubsection{Stage 2: Skill-Exploitation Fine-Tuning without Harness}

Building on Stage~1, Stage~2 enables the model to perform task-relevant actions by fine-tuning the hypernetwork on skill-exploitation trajectories without a coding-agent harness. Each training instance pairs a textual skill \(s\) with a supervised single- or multi-turn trajectory \(\tau\) that demonstrates its use in a concrete task. The textual skill conditions parametric skill generation, while the adapted language model \(\mathcal{M}_{\Phi(s)}\) learns to produce responses given the task context without the textual skill. We optimize the hypernetwork over the supervised tokens:
\begin{equation}
  \mathcal{L}_{\mathrm{exploit}}(\theta)
  =
  -\sum_{t=1}^{T}
  a(t)
  \log
  \mathcal{M}_{\Phi(s)}
  \bigl(x_t \mid x_{<t}\bigr),
\end{equation}
where \(\tau=(x_1,\ldots,x_T)\) is the serialized trajectory and \(a(t)\in\{0,1\}\) is a supervision mask such that \(a(t)=1\) if the \(t\)-th token is a model output to be supervised and \(a(t)=0\) otherwise.

\subsubsection{Stage 3: Skill-Exploitation Fine-Tuning with Harness}

Stage 3 further enables the model to exploit skills in realistic agentic settings. The training trajectories in this stage are substantially more challenging: they involve long-horizon planning, context inspection, tool invocation, code modification, and verification, with the model coordinating its actions against the evolving task state and environment feedback. We use the same \(\mathcal{L}_{\mathrm{exploit}}\) objective as in Stage 2. Full training settings are provided in Appendix~\ref{app:training-setup}.

\subsection{Training Data Curation of \textsc{ParametricSkills}}

Figure~\ref{fig:skills} presents the statistics of the training dataset for each designed training stage. Specifically, we construct the training data for \textsc{ParametricSkills} through the following pipeline: (1) collecting, filtering, and normalizing reusable textual skills; (2) expanding collected textual skills into skill-exploitation trajectories without harness; and (3) further utilizing the skill-exploitation trajectories with harness constructed by \citet{tan2026skt}. 

\subsubsection{Collection of Textual Skills}
We curate textual skills from two complementary sources. For publicly available textual skills, we crawl candidate textual skills from online sources, including skills.sh, SkillsMP, skilldb, npm, and GitHub. We then normalize the collected skills and filter out invalid or uninformative ones. For experience-driven skills, we extract experience-driven skills from high-quality agentic trajectories, grounding each skill in the underlying execution evidence. To transform them into reusable textual skills, we use a three-stage pipeline of short model calls. (1) The summarize stage compresses each preprocessed trajectory into a structured JSON summary that captures key evidence, localization steps, modification strategies, and verification signals. (2) The extract stage generates multiple candidate skills from the summary and selects the strongest candidate based on reusability and non-triviality. (3) Finally, the validate stage checks whether the candidate is supported by the trajectory and assesses its actionability, potential for transfer across repositories, non-triviality, and value as a standalone training document. After filtering, 101{,}119 skills remain in the training split of the pretraining corpus. 

\subsubsection{Construction of Skill-Exploitation Trajectories}
After skill pretraining corpus curation, we construct single- or multi-turn skill-exploitation trajectories without a coding-agent harness for the curated skills, where the model must correctly use the skills provided with related tasks. 
Single-turn examples pair a skill with a user request and the teacher-generated reasoning and solution. In each trajectory, the teacher model is required to use a specific part of the skill or combine multiple parts of the same skill to derive a solution. The model learns to apply the skill's procedures and behavioral requirements to the task's needs, follow user constraints, and produce a concrete solution or artifact. Multi-turn examples pair a task request and interaction history with a target response or action for the current turn. The history contains prior responses or actions and the observations or feedback available so far. Such tasks require multiple turns because later decisions depend on earlier interactions, and each action can change the task state. The model learns to track progress, interpret feedback in light of prior actions, and determine the next step while adhering to the skill's procedures and behavioral requirements.



\subsection{Self-Evolving and Continual Learning}
\label{subsec:method-skill-evolution}

\subsubsection{Self-Evolving}

Beyond skill parameterization, we introduce a self-evolving parametric skill loop, where skills are refined in text space and verified in parameter space through the following steps: (1) initial skill generation: given a task instruction, if an initial skill does not exist, a skill generator produces the initial skill, which is then transformed into a parametric skill by the hypernetwork. (2) verification: the model generates a solution using the current parametric skill, after which a verifier assesses whether the solution meets the task requirements and returns feedback. (3) revision: if the candidate is rejected, the feedback is used to revise the textual skill. The revised skill is recompiled into a parametric skill for the next iteration. (4) termination check: the loop runs for a fixed number of rounds or until verifier acceptance.

\subsubsection{Continual Learning}

We further establish a continual learning framework, in which the self-evolving loop incrementally builds a global parametric skill by merging skills acquired from newly solved tasks. For each task, a base parametric skill is generated, while inference additionally draws a lightweight contribution from the current global parametric skill. If the solution to the current task is accepted, the corresponding trajectory is summarized as a skill and transformed into a parametric skill, normalized by its effective update magnitude, and merged into the global parametric skill via an exponential moving average (EMA; Appendix~\ref{app:online-continual-details}). The global parametric skill thus serves as a reusable prior distilled from previously successful trajectories, enabling test-time continual learning.
\section{Experiments}
\label{sec:experiments}

We conduct separate evaluations on the three training stages of our pipeline. Specifically, we first evaluate the effectiveness of Stage~1 training by comparing the reconstruction performance on held-out textual skills. We then evaluate Stage~2 training on ALFWorld~\citep{shridhar2020alfworld}, which comprises six types of interactive household tasks.
Finally, we evaluate the final checkpoints of \textsc{ParametricSkills} on SkillEval~\citep{tan2026skt}, which evaluates skill use in agentic task execution, containing single-skill, two-skill, and three-skill scenarios.

\paragraph{Implementation.}
We implement \textsc{ParametricSkills} on three distinct backbones, Qwen3-8B, Qwen3.5-4B, and Qwen3.5-9B, to validate its effectiveness across model architectures and sizes. We follow the architecture of SHINE~\citep{liu2026shine} to implement the hypernetworks on the three backbones.
All hypernetworks are first trained with the pretraining process of SHINE (denoted as SHINE in our comparisons), and then trained on our three-stage pipeline.
Implementation details and evaluation configurations are provided in Appendix~\ref{app:training-setup} and Appendix~\ref{app:evaluation-details}.

\paragraph{Baselines.}
We evaluate four baselines on both ALFWorld and SkillEval: \textbf{SFT (LoRA)}, \textbf{No-Skill}, \textbf{SHINE}, and \textbf{ICL}. SFT (LoRA) trains LoRA adapters~\citep{hu2021lora} through supervised fine-tuning.
No-Skill runs the model without any skill input,
while ICL runs with the in-context skill.

\subsection{Main Experiments}

\subsubsection{Evaluation of Stage~1 Training}

\newcommand{\stageonefigure}[1]{%
    \centering
    \includegraphics[
        width=\linewidth
    ]{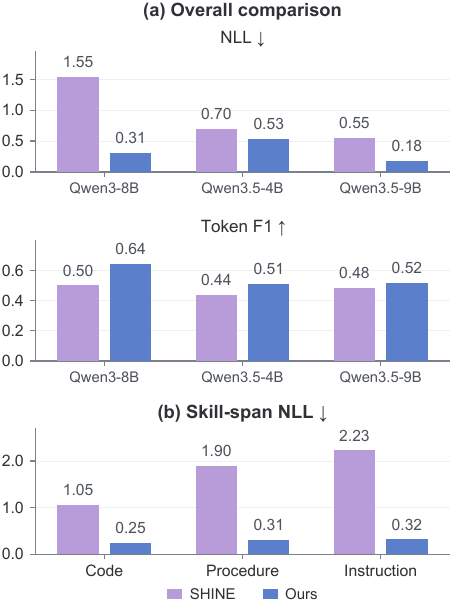}
    \caption{#1}
    \label{fig:pretraining-comparison}}
\ifdefined\TechReport
\begin{wrapfigure}{r}{0.5\columnwidth}
\vspace{-14pt}
\captionsetup{skip=3pt}
\stageonefigure{Stage~1 reconstruction results.}
\vspace{-14pt}
\end{wrapfigure}
\fi

Figure~\ref{fig:pretraining-comparison} reveals that SHINE's general pretraining is insufficient for faithful skill-content internalization. Our Stage~1 training consistently improves reconstruction across all three backbones, while also improving token-level F1. Notably, reconstruction improves for critical spans such as code, procedures, and instructions, indicating more faithful preservation of skill-specific details in the generated parameters. These results establish a stronger foundation for subsequent skill exploitation and agentic training.

\ifdefined\TechReport\else
\begingroup
\floatsNormal
\begin{figure}[t]
\stageonefigure{
        Stage~1 reconstruction results:
        (a) overall NLL and token-level F1;
        (b) span-level reconstruction NLL.
    }
\end{figure}
\endgroup
\fi

\subsubsection{Evaluation of Stage~2 Training}
\label{subsec:alfworld-main}
\begin{table*}[t]
\centering
\caption{ALFWorld results on Qwen3-8B, Qwen3.5-4B, and Qwen3.5-9B.
The highest results within each backbone, including ties, are bolded;
dashes indicate unreported results.}
\label{tab:alfworld-main}

\normalsize
\setlength{\tabcolsep}{0pt}
\renewcommand{\arraystretch}{1.08}

\scalebox{0.9}{%
\begin{tabular}{@{}>{\raggedright\arraybackslash}p{0.24\textwidth}*{7}{>{\centering\arraybackslash}p{\dimexpr0.76\textwidth/7\relax}}@{}}
\toprule
\shortstack[l]{\textbf{Method}\\\phantom{(274)}}
& \shortstack{\textbf{Pick}\\(59)}
& \shortstack{\textbf{Look}\\(31)}
& \shortstack{\textbf{Clean}\\(58)}
& \shortstack{\textbf{Heat}\\(39)}
& \shortstack{\textbf{Cool}\\(46)}
& \shortstack{\textbf{Pick2}\\(41)}
& \shortstack{\textbf{Overall$\uparrow$}\\(274)} \\
\midrule

\rowcolor{gray!9}
\multicolumn{8}{l}{\textit{Qwen3-8B}} \\

\rowcolor{contextblue!60}
SFT (LoRA)
& 94.92
& 87.10
& 93.10
& 84.62
& \textbf{91.30}
& \textbf{92.68}
& 91.24 \\

No-Skill
& 98.31 & 51.61 & 48.28 & 25.64 & 32.61 & 68.29 & 56.57 \\

SHINE
& 40.68 & 9.68 & 15.52 & 12.82 & 10.87 & 12.20 & 18.61 \\

ICL
& 94.92 & 74.19 & 82.76 & 82.05 & 50.00 & 63.41 & 75.91 \\

\rowcolor{skillpurple!45}
\textsc{ParametricSkills}
& \textbf{100.00}
& \textbf{90.32}
& \textbf{96.55}
& \textbf{87.18}
& 89.13
& \textbf{92.68}
& \textbf{93.43} \\

\addlinespace[0.2em]
\rowcolor{gray!9}
\multicolumn{8}{l}{\textit{Qwen3.5-4B}} \\

\rowcolor{contextblue!60}
SFT (LoRA)
& 84.75
& \textbf{93.55}
& 82.76
& \textbf{92.31}
& 84.78
& 56.10
& 82.12 \\

No-Skill
& 55.93
& 9.68
& 17.24
& 23.08
& 15.22
& 26.83
& 26.64 \\

SHINE
& 50.85 & 3.23 & 1.72 & 5.13 & 4.35 & 31.71 & 17.88 \\

ICL
& 22.03
& 48.39
& 29.31
& 17.95
& 8.70
& 26.83
& 24.45 \\

\rowcolor{skillpurple!45}
\textsc{ParametricSkills}
& \textbf{93.22}
& 90.32
& \textbf{91.38}
& \textbf{92.31}
& \textbf{86.96}
& \textbf{85.37}
& \textbf{90.15} \\

\addlinespace[0.2em]
\rowcolor{gray!9}
\multicolumn{8}{l}{\textit{Qwen3.5-9B}} \\

\rowcolor{contextblue!60}
SFT (LoRA)
& \textbf{98.31}
& 96.77
& 93.10
& \textbf{97.44}
& 93.48
& \textbf{87.80}
& 94.53 \\

No-Skill
& 96.61
& 51.61
& 39.66
& 56.41
& 19.57
& 73.17
& 57.30 \\

SHINE
& 84.75 & 32.26 & 51.72 & 23.08 & 6.52 & 34.15 & 42.34 \\

ICL
& 77.97
& 54.84
& 56.90
& 61.54
& 52.17
& 73.17
& 63.50 \\

\rowcolor{skillpurple!45}
\textsc{ParametricSkills}
& 96.61
& \textbf{100.00}
& \textbf{98.28}
& \textbf{97.44}
& \textbf{95.65}
& 82.93
& \textbf{95.26} \\

\bottomrule
\end{tabular}
}
\end{table*}

As shown in Table~\ref{tab:alfworld-main}, we evaluate the Stage~2 checkpoints of \textsc{ParametricSkills} on ALFWorld. \textsc{ParametricSkills} achieves the best average performance on all three backbones. Specifically, on Qwen3-8B, it obtains an average success rate of 93.43\%, improving over ICL prompting by 17.52 points. On Qwen3.5-4B, \textsc{ParametricSkills} reaches 90.15\%, surpassing ICL by 65.70 points; notably, the ICL baseline even underperforms No-Skill on average, indicating that smaller models struggle to exploit textual skills through ICL. On Qwen3.5-9B, \textsc{ParametricSkills} achieves 95.26\%, exceeding ICL by 31.76 points. Compared with SHINE, the overall success rates improve by 74.82, 72.27, and 52.92 percentage points on Qwen3-8B, Qwen3.5-4B, and Qwen3.5-9B, respectively. Across individual task categories, \textsc{ParametricSkills} attains the highest or joint-highest score in 14 of the 18 backbone and task combinations, with especially large gains on categories where both No-Skill and ICL are weak. 
These results demonstrate that \textsc{ParametricSkills} effectively parameterizes the skill-exploitation methodology after Stage~2 training.

\subsubsection{Evaluation of Stage~3 Training}

As shown in Table~\ref{tab:skilleval-main}, we evaluate the final checkpoint of \textsc{ParametricSkills} and the baselines on SkillEval to assess skill exploitation in real-world agentic scenarios.
On Qwen3-8B, \textsc{ParametricSkills} scores 48.89, exceeding SHINE (18.05),
ICL, and No-Skill by 30.84, 6.41, and 5.32 points, respectively.
It ranks second overall and in K1/K2, behind SFT (LoRA), but achieves the
highest K3 mean (58.06), 12.84 points above ICL. On Qwen3.5-4B, \textsc{ParametricSkills} scores 57.83, exceeding SHINE (13.53) by 44.30 points, while remaining close to ICL (57.79)
and below No-Skill (58.56) and SFT (LoRA) (61.42). On Qwen3.5-9B, it achieves the highest reported means overall and in K1/K2/K3. Its overall score of 63.12 exceeds SHINE by 46.74
points and ICL, No-Skill, and SFT (LoRA) by 2.16, 2.89, and 1.14 points, respectively. The consistent gains over SHINE align with our design of reconstructing skill content for internalization and supervising its use in tasks and agent interactions. These results support applicability across backbones, with clearer gains over ICL and SFT (LoRA) on Qwen3.5-9B and limited benefit on Qwen3.5-4B.

\begin{table}[htbp]
\centering
\caption{
SkillEval results on Qwen3-8B, Qwen3.5-4B, and Qwen3.5-9B.
Scores are mean $\pm$ standard deviation over four independent runs.
Within each backbone, the highest and second-highest means are
bolded and underlined.
}
\label{tab:skilleval-main}

\normalsize
\setlength{\tabcolsep}{0pt}
\renewcommand{\arraystretch}{1.08}

\scalebox{0.9}{%
\begin{tabular}{
@{}
>{\raggedright\arraybackslash}p{0.24\linewidth}
*{4}{>{\centering\arraybackslash}p{0.19\linewidth}}
@{}
}
\toprule
\textbf{Method}
& \textbf{K1}
& \textbf{K2}
& \textbf{K3}
& \textbf{Overall} \\
\midrule

\rowcolor{gray!9}
\multicolumn{5}{l}{\textit{Qwen3-8B }} \\

\rowcolor{contextblue!60}
SFT (LoRA)
& \best{56.17 $\pm$ 3.52}
& \best{49.33 $\pm$ 1.11}
& \second{56.07 $\pm$ 5.72} 
& \best{53.00 $\pm$ 2.17} \\

No-Skill
& 41.53 $\pm$ 6.13
& 39.75 $\pm$ 4.96
& 53.46 $\pm$ 4.81 
& 43.57 $\pm$ 3.15 \\

SHINE
& 16.43 $\pm$ 2.63
& 17.43 $\pm$ 4.51
& 21.26 $\pm$ 2.03
& 18.05 $\pm$ 2.87\\

ICL

& 40.54 $\pm$ 4.68
& 42.31 $\pm$ 4.57
& 45.22 $\pm$ 3.78 
& 42.48 $\pm$ 2.69 \\

\rowcolor{skillpurple!45}
\textsc{ParametricSkills}
& \second{48.88 $\pm$ 5.29}
& \second{44.11 $\pm$ 2.70}
& \best{58.06 $\pm$ 3.73}
& \second{48.89 $\pm$ 1.73} \\

\addlinespace[0.2em]
\rowcolor{gray!9}
\multicolumn{5}{l}{\textit{Qwen3.5-4B }} \\

\rowcolor{contextblue!60}
SFT (LoRA)
& \best{63.96 $\pm$ 5.96}
& \best{57.93 $\pm$ 1.14}
& \second{64.92 $\pm$ 2.05}
& \best{61.42 $\pm$ 2.06} \\

No-Skill
& 57.33 $\pm$ 3.52
& \second{55.58 $\pm$ 1.74}
& \best{65.82 $\pm$ 2.69}
& \second{58.56 $\pm$ 1.33} \\

SHINE
& 12.81 $\pm$ 6.17
& 12.62 $\pm$ 3.53
& 16.16 $\pm$ 6.48
& 13.53 $\pm$ 1.34 \\

ICL
& \second{58.14 $\pm$ 1.98}
& 54.32 $\pm$ 1.73
& 63.98 $\pm$ 2.54
& 57.79 $\pm$ 0.86 \\

\rowcolor{skillpurple!45}
\textsc{ParametricSkills}
& 57.72 $\pm$ 2.63
& 55.10 $\pm$ 2.21
& 63.19 $\pm$ 3.17
& 57.83 $\pm$ 1.49 \\

\addlinespace[0.2em]
\rowcolor{gray!9}
\multicolumn{5}{l}{\textit{Qwen3.5-9B }} \\

\rowcolor{contextblue!60}
SFT (LoRA)
& 60.71 $\pm$ 2.14
& 59.28 $\pm$ 1.00
& \second{68.75 $\pm$ 3.27}
& \second{61.98 $\pm$ 1.28} \\

No-Skill
& 56.07 $\pm$ 0.88
& \second{59.49 $\pm$ 2.17}
& 66.84 $\pm$ 1.69
& 60.23 $\pm$ 0.88 \\

SHINE
& 20.47 $\pm$ 5.69
& 14.91 $\pm$ 3.23
& 14.09 $\pm$ 4.71
& 16.38 $\pm$ 1.62 \\

ICL
& \second{61.56 $\pm$ 3.01}
& 57.11 $\pm$ 0.72
& 67.59 $\pm$ 2.56
& 60.96 $\pm$ 1.08 \\

\rowcolor{skillpurple!45}
\textsc{ParametricSkills}
& \best{61.82 $\pm$ 1.61}
& \best{60.52 $\pm$ 0.67}
& \best{69.73 $\pm$ 5.78}
& \best{63.12 $\pm$ 1.74} \\

\bottomrule
\end{tabular}
}
\end{table}

\subsection{Self-Evolving and Continual Learning}
\label{subsec:skilleval-self-evolution}

We separately explore accumulation across tasks through a global parametric skill updated in parameter space. In an online sequence of tasks, the self-evolving loop incrementally builds a global parametric skill by merging skills acquired from newly solved tasks. For each task, a base parametric skill is still generated, while inference additionally draws a lightweight contribution from the current global parametric skill. If the solution to the current task is accepted, the final task-specific parametric skill is normalized by its effective update magnitude and merged into the global parametric skill via an exponential moving average (EMA; Appendix~\ref{app:online-continual-details}). The global parametric skill thus serves as a reusable prior accumulated from previously accepted task adapters, equipping the model for upcoming tasks and enabling test-time continual learning without backpropagation and with negligible computational and memory overhead.

\begingroup
\ifdefined\origtable
  \let\table\origtable
  \let\endtable\endorigtable
\fi
\begin{table}[htbp]
\centering
\captionsetup{font=small}
\caption{Self-evolution results on the single-skill subset of SkillEval with Qwen3-8B.}
\label{tab:skilleval-self-evolution-main}
\small
\setlength{\tabcolsep}{0pt}
\renewcommand{\arraystretch}{1.08}
\begin{tabular}{
@{}
>{\raggedright\arraybackslash}p{0.50\columnwidth}
>{\centering\arraybackslash}p{0.32\columnwidth}
@{}
}
\toprule
\textbf{Method} & \textbf{K1} \\
\midrule
No evolution & 46.89 $\pm$ 1.09 \\
Self-evolution & 53.18 $\pm$ 1.30 \\
Continual learning & \textbf{56.47} $\pm$ 3.05 \\
\bottomrule
\end{tabular}
\end{table}
\endgroup

\section{Discussion}
\label{sec:discussion}

\subsection{Effect of Multi-Skill Composition Strategies}
\label{subsec:exp-multi-lora-merging}

We evaluate multi-skill composition strategies on the multi-skill tasks in SkillEval. The experiment is conducted using Qwen3-8B and the Stage~3 checkpoint, with all other settings identical to those of the main experiment on SkillEval. We compare three strategies: (1) Text concatenation: concatenating the textual skills and converting them into one parametric skill. (2) Linear merge: compiling each skill independently and averaging the resulting parametric skills. (3) Rank-concat merge: concatenating the factors of independently generated parametric skills along their rank dimensions. Table~\ref{tab:multi-lora-merging} reports the overall scores. Text concatenation achieves the highest score among the three strategies on both subsets. Its overall score is 50.93, exceeding linear merge by 6.65 points and rank-concat merge by 3.91 points. We therefore adopt text concatenation followed by compilation into one LoRA as our multi-skill composition strategy.

\begingroup
\ifdefined\origtable
  \let\table\origtable
  \let\endtable\endorigtable
\fi
\begin{table}[htbp]
\centering
\captionsetup{font=small}
\caption{Multi-skill composition on SkillEval with Qwen3-8B.}
\label{tab:multi-lora-merging}
\small
\setlength{\tabcolsep}{0pt}
\renewcommand{\arraystretch}{1.08}

\begin{tabular}{
@{}
>{\raggedright\arraybackslash}p{0.52\columnwidth}
*{3}{>{\centering\arraybackslash}p{0.16\columnwidth}}
@{}
}
\toprule
\textbf{Method}
& \textbf{K2}
& \textbf{K3}
& \textbf{Overall} \\
\midrule
\rowcolor{skillpurple!45}
Linear merge & 41.53 & 49.55 & 44.28 \\
Rank-concat merge & 42.72 & 55.23 & 47.01 \\
Text concatenation & \textbf{47.32} & \textbf{57.85} & \textbf{50.93} \\
\bottomrule
\end{tabular}
\end{table}
\endgroup

\subsection{Effect of Training Stages}

We ablate the effect of training stages on Qwen3.5-9B using SkillEval
(Table~\ref{tab:training-stage-ablation}). The full pipeline achieves the highest overall mean, 63.12. Omitting Stages~1, 2, and 3 lowers this score by 3.26, 12.99,
and 9.14 points, respectively. The largest drop occurs without Stage~2, supporting Stage~2 training as a bridge between skill-content parameterization
and complex skill exploitation.

\begingroup
\ifdefined\origtable
  \let\table\origtable
  \let\endtable\endorigtable
\fi

\begin{table}[htbp]
\centering
\captionsetup{font=small}
\caption{Training-stage ablation on SkillEval with Qwen3.5-9B.}
\label{tab:training-stage-ablation}
\small
\setlength{\tabcolsep}{0pt}
\renewcommand{\arraystretch}{1.08}

\begin{tabular}{
@{}
>{\raggedright\arraybackslash}p{0.30\columnwidth}
>{\centering\arraybackslash}p{0.40\columnwidth}
@{}
}
\toprule
\textbf{Pipeline} & \textbf{Overall} \\
\midrule
W/o Stage~1 & 59.86 $\pm$ 0.97 \\
W/o Stage~2 & 50.13 $\pm$ 1.74 \\
W/o Stage~3 & 53.98 $\pm$ 3.97 \\
\rowcolor{skillpurple!45}
Full pipeline & \best{63.12 $\pm$ 1.74} \\
\bottomrule
\end{tabular}
\end{table}
\endgroup

\section{Related Works}
\paragraph{Test-time parameter generation.}
Generating adapter weights on the fly rather than fine-tuning has gained traction as a test-time adaptation paradigm. Hypernetwork-based~\citep{ha2016hypernetworks} approaches map task descriptions or task embeddings directly into LoRA parameters. For instance, Text-to-LoRA~\citep{charakorn2025text} generates task-oriented LoRA adapters by inputting the task description embeddings into an MLP-based hypernetwork with a single forward pass to avoid costly task-specific fine-tuning of LLMs; 
LoRA-Gen~\citep{xiao2025lora-gen} uses a cloud-side language
model to condition a layer-wise mixture of LoRA experts on
task descriptions, generating task-specific adapters for
smaller edge-side models.
Rather than generating parameters from specific task descriptions, Doc-to-LoRA~\citep{charakorn2026doc} generates adapters from any unseen text to enable subsequent question-answering without the explicit context; while SHINE~\citep{liu2026shine} concurrently extends this idea by utilizing the parameters of the LLM to construct the hypernetwork, thereby reusing the powerful in-context understanding capabilities of the LLM. In vision, LoRA.rar~\citep{shenaj2025lora.rar} uses a hypernetwork to predict merging coefficients for composing two LoRAs.

\paragraph{Skill exploitation and self-evolution.} Representing agentic experience as reusable skills, i.e., textual recipes that encode proven problem-solving strategies, has become a standard ingredient in modern coding agents. Recent work focuses on evolving such skills over the agent's lifetime: CoEvoSkills~\citep{zhang2026coevoskills} couples a skill generator with a co-evolving verifier through generate-verify-refine cycles, reporting 46\% improvements; SkillOpt~\citep{yang2026skillopt}
treats natural-language skill documents as trainable parameters and optimizes them via trajectory-driven text edits, achieving 52/52 wins across benchmarks; OpenSpace~\citep{OpenSpace2026} provides an open-source self-evolving skill engine that captures patterns from agent interactions.
SE-Agent~\citep{lin2025se} optimizes reasoning trajectories rather than skills directly. All of these methods operate in text space, relying on LLM-based rewriting that is inherently noisy, drifts over iterations, and consumes context tokens at inference. Moreover, prior work has shown that
in-context skill exploitation is bottlenecked by the model's instruction-following capability, with smaller models failing to act on complex skill descriptions even when they possess the underlying competence. 
\textsc{ParametricSkills} complements text-space evolution by
compiling refined skills into parametric adapters, supporting
multi-skill use and verifier-guided skill accumulation without
retaining skill documents in the task model's context.

\section{Conclusion}
We present \textsc{ParametricSkills}, a framework that converts
free-form textual skills into LoRA adapters at test time through
a single hypernetwork forward pass.
These adapters provide reusable skills without
keeping skill documents in the context.
Experiments on ALFWorld and SkillEval show improvements
over ICL across three backbones.
We further demonstrate task-level skill self-evolution and explore
online accumulation of skills from successful tasks into a global
parametric skill.
These initial results suggest a promising route toward test-time
continual learning through skill refinement, recompilation, and
parameter accumulation without backpropagation.
\section*{Limitations}

\paragraph{Backbone-specific training.}
Our framework requires training a separate hypernetwork for each backbone, introducing additional training costs. However, our primary goal is to establish the feasibility
of hypernetwork-generated parametric skills and explore their potential for skill-exploitation and continual learning, rather than to optimize training efficiency. We view hypernetwork training as a potential component of LLM post-training, whose additional cost may be acceptable compared to other post-training processes such as reinforcement learning.

\paragraph{Skill switching and cache reuse.}
Switching parametric skills by replacing LoRA adapters can invalidate previously computed KV caches. However, cache invalidation is also a challenge for in-context skill management in existing agent frameworks. Even when textual skills are appended to the conversation, replacing or removing a skill from an earlier turn invalidates subsequent cached states, with the extent depending on its position in the context. Thus, both representations face overhead in dynamic skill switching, albeit to different degrees. Developing more efficient switching and cache-reuse mechanisms remains an important direction for future work.

\paragraph{Continual learning at scale.}
Although our preliminary experiments demonstrate the potential of parametric skills for self-evolution and continual learning, they are limited to small-scale verification. Given the complexity and importance of continual learning, larger-scale studies are needed to investigate long-term knowledge retention, interference between accumulated skills, and robustness to imperfect feedback. Exploring more effective mechanisms for continual accumulation remains a key direction for future work.

\paragraph{Evaluation scope.}
Our evaluation covers three Qwen backbones and two downstream benchmarks, ALFWorld and SkillEval. Generalization to other model families, task domains, and larger or potentially conflicting skill sets remains to be established.

\bibliographystyle{plainnat}
\bibliography{colm2026_conference}

\clearpage
\trappendix
\ifdefined\TechReport\else

\fi
\section{Comparison to LatentSkill}
Concurrent with our work, LatentSkill~\citep{yu2026latentskill} applies text-to-LoRA generation to convert agentic skills into plug-and-play adapters, demonstrating gains
over in-context skills on the embodied agent benchmark ALFWorld~\citep{shridhar2020alfworld} and the search-augmented question-answering benchmark Search-QA~\citep{jin2025search-r1}. \textsc{ParametricSkills} shares the skill-to-LoRA backbone but differs in four aspects: (i) we target real-world agentic tasks such as production-level software engineering rather than a simple demonstration on ALFWorld and Search-QA, (ii) we construct a comprehensive skill library containing skills summarized from real-world agentic trajectories, such as those collected from OpenCode, a popular vibe-coding framework, and we further construct a massive number of trajectories that use these skills to solve problems using a strong teacher model; (iii) we introduce principled multi-skill composition targeting real-world problems that frequently require more than two skills to solve, and (iv) we build a self-evolving parametric skill pipeline and show that parametric skills can be a gateway for continual learning of LLMs, which LatentSkill does not address.
\section{Training Setup}
\label{app:training-setup}

All stages use our PyTorch and Hugging Face Transformers implementation with AdamW, BF16 precision, gradient checkpointing, and a warmup ratio of 0.03. Textual skill inputs are truncated at 12,288 tokens throughout training. The maximum training length is set to 12,352 tokens in Stage~1 and 24,576 tokens in Stages~2 and~3. For the reweighting loss, we use $\beta=0.25$ and normalized category weights proportional to $1:2:3$ for code, procedure, and instruction spans, respectively. Loss aggregation and weighting experiments are detailed in Appendix~\ref{app:reconstruction-ablation}. Table~\ref{tab:pretraining-config} summarizes the schedules and effective global batch sizes for each backbone model.

\begin{table*}[htbp]
\centering
\caption{Training configurations by backbone and stage. Batch sizes are effective global batch sizes. }
\label{tab:pretraining-config}
\begingroup
\small
\setlength{\tabcolsep}{7pt}
\renewcommand{\arraystretch}{1.1}
\begin{tabular}{@{}lcccc@{}}
\toprule
Stage & Learning rate & Batch size & Configured epochs & Scheduler \\
\midrule
\multicolumn{5}{l}{\textit{Qwen3-8B}} \\
Stage~1 & $1\times10^{-5}$ & 32 & 2 & Linear \\
Stage~2 & $1\times10^{-5}$ & 8 & 1 & Linear \\
Stage~3 & $3\times10^{-6}$ & 32 & 2 & Linear \\
\midrule
\multicolumn{5}{l}{\textit{Qwen3.5-4B and Qwen3.5-9B}} \\
Stage~1 & $1\times10^{-5}$ & 32 & 2 & Linear \\
Stage~2 & $2\times10^{-5}$ & 2 & 1 & Linear \\
Stage~3 & $5\times10^{-5}$ & 8 & 2 & Cosine \\
\bottomrule
\end{tabular}
\endgroup
\end{table*}

\section{Continual Learning Details}
\label{app:online-continual-details}

This section describes the implementation details of the continual learning experiment. The goal is to maintain a global parametric skill that stores accumulated task experience while still allowing each new task to use its own task-specific adapter.

Let \(G_{t-1}\) denote the global parametric skill before processing task \(t\), and let
\(\Delta_t^{(r)}\) denote the task-specific parametric skill produced at evolution round \(r\) for task \(t\). At each round, the model predicts with the current task-specific parametric skill combined with the contribution from the current global parametric skill:
\[
    \Delta_{\mathrm{infer}}^{(r)}
    =
    \Delta_t^{(r)} + \lambda G_{t-1},
\]
where \(\lambda\) controls the strength of the accumulated global parametric skill. In our experiment, we set \(\lambda=0.05\).

After prediction, the verifier checks the candidate answer. If the candidate is rejected and the round budget has not been exhausted, the verifier feedback is used to revise the skill document, producing a new task-specific parametric skill for the next round. Let \(\Delta_t^\star\) denote the final parametric skill when the task-level
self-evolution loop stops.

If the final candidate is accepted, we treat \(\Delta_t^\star\) as useful task experience and merge it into the global parametric skill. Before the merge, we normalize the final parametric skill by the norm of its effective update:
\[
    \bar{\Delta}_t
    =
    \frac{\Delta_t^\star}{\|\Delta_t^\star\|_F}.
\]
Here, \(\|\Delta_t^\star\|_F\) denotes the Frobenius norm of the effective LoRA update induced by the parametric skill. This normalization prevents a high-norm parametric skill from dominating the accumulated global parametric skill only because of scale.

The global parametric skill is then updated with an exponential moving average (EMA):
\[
    G_t = (1-\beta)G_{t-1} + \beta \bar{\Delta}_t,
\]
where \(\beta\) is the EMA update coefficient. In our experiment, we set
\(\beta=0.1\), so each accepted parametric skill contributes 10\% to the updated global parametric skill. If the final candidate is rejected, the global parametric skill is left unchanged:
\[
    G_t = G_{t-1}.
\]

Thus, accepted task-specific parametric skills are accumulated as a reusable LoRA-space memory, while rejected task adapters are discarded. 

\section{Training Curves}

\begin{figure*}[htbp]
\centering
\includegraphics[width=\linewidth]{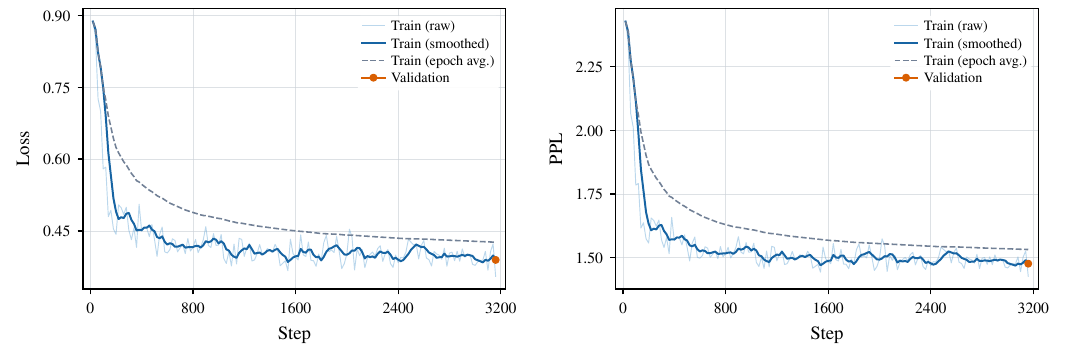}
\caption{Loss and PPL Curves of the Skill Reconstruction Pretraining Stage}
\label{fig:pretrain-loss}
\vspace{-3mm}
\end{figure*}

\begin{figure*}[htbp]
\centering
\includegraphics[width=\linewidth]{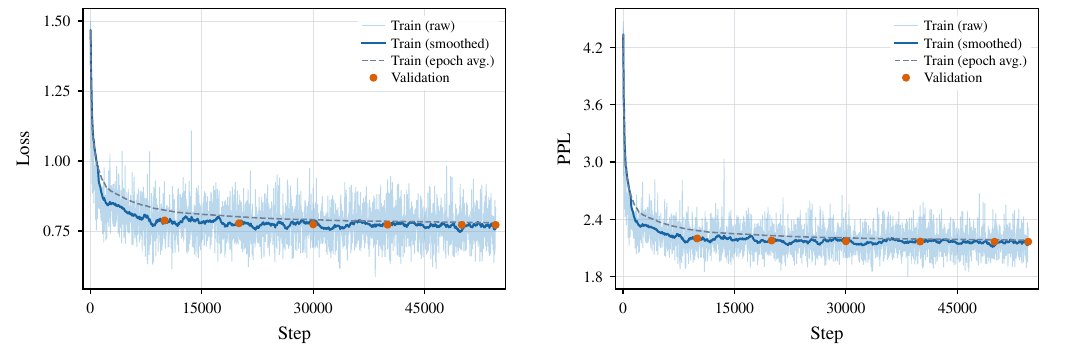}
\caption{Loss and PPL Curves of the Skill Exploitation Fine-Tuning Stage}
\label{fig:stage2-loss}
\vspace{-3mm}
\end{figure*}

\begin{figure*}[htbp]
\centering
\includegraphics[width=\linewidth]{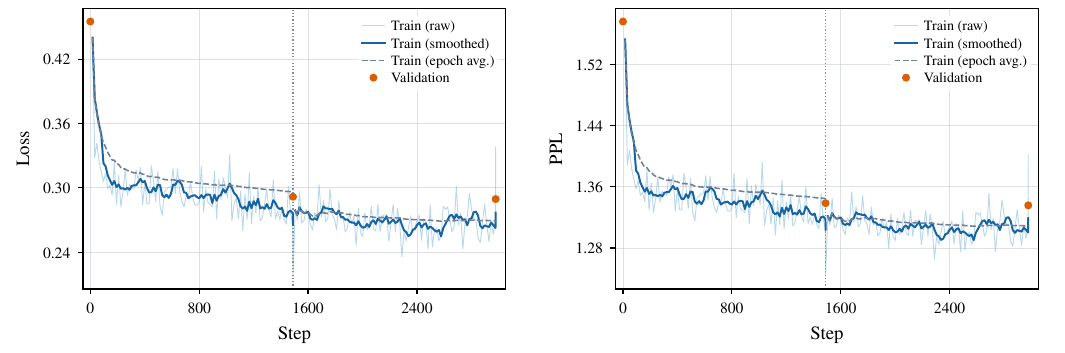}
\caption{Loss and PPL Curves of the Agentic Fine-Tuning Stage}
\label{fig:stage3-loss}
\vspace{-3mm}
\end{figure*}

We provide the training curves of the three stages on Qwen3.5-9B in Figures~\ref{fig:pretrain-loss}--\ref{fig:stage3-loss}. Across all three stages, training loss and perplexity decrease markedly and settle into stable trends, indicating effective optimization of each stage's objective.
Stage~1 rapidly reduces the reconstruction loss and reaches a stable low-loss trend, reflecting effective learning of the skill reconstruction objective. In Stage~2, substantial training-loss reduction is accompanied by consistent improvements in validation loss and perplexity, supporting successful fine-tuning for single- and multi-turn skill exploitation. Stage~3 produces a clear improvement on the held-out reasoning objective: validation loss decreases from~0.4551 to~0.2895 and perplexity from~1.5763 to~1.3358, with additional gains in the second epoch.

\newcommand{\weightingablationcontent}{%
\centering
\caption{Ablation of the auxiliary reweighting loss in skill reconstruction.
$\beta=0$ removes the auxiliary term; ratios are ordered as
code:procedure:instruction.
Best scores and lowest latency are bolded.}
\label{tab:weighting-ablation}
\begingroup
\small
\setlength{\tabcolsep}{0pt}
\renewcommand{\arraystretch}{1.15}

\begin{tabular}{@{}
>{\centering\arraybackslash}p{0.14\textwidth}
>{\centering\arraybackslash}p{0.07\textwidth}
*{5}{>{\centering\arraybackslash}p{0.158\textwidth}}
@{}}
\toprule
\multicolumn{1}{c}{Weight ratio}
& \multicolumn{1}{c}{$\beta$}
& \multicolumn{1}{c}{Micro F1 $\uparrow$}
& \multicolumn{1}{c}{\shortstack{Skill-macro\\F1 $\uparrow$}}
& \multicolumn{1}{c}{\shortstack{Avg. generated\\tokens}}
& \multicolumn{1}{c}{\shortstack{LLM judge\\(0--100) $\uparrow$}}
& \multicolumn{1}{c}{\shortstack{Avg. latency\\(s) $\downarrow$}}
\\
\midrule
\textemdash
& $0$ & 0.3914 & 0.3931 & 910 & 38.38 & \textbf{15.24} \\
\texttt{1:2:3}
& $0.10$ & \textbf{0.3944} & \textbf{0.3953} & 1049.78 & 39.05 & 19.49 \\
\midrule
\texttt{1:1:1}
& $0.25$ & 0.3770 & 0.3798 & 1106.2 & 39.37 & 26.91 \\
\rowcolor{skillpurple}
\texttt{1:2:3}
& $0.25$ & 0.3723 & 0.3754 & 1166.3 & \textbf{39.73} & 27.94 \\
\texttt{1:2:5}
& $0.25$ & 0.3627 & 0.3656 & 1295.8 & 38.96 & 31.89 \\
\texttt{1:3:5}
& $0.25$ & 0.3706 & 0.3709 & 1169.6 & 37.68 & 28.44 \\
\bottomrule
\end{tabular}

\endgroup
}
\newcommand{\weightingablationtable}{%
\ifdim\columnwidth<\textwidth
\twocolumn[{%
\begin{minipage}{\textwidth}
\captionsetup{type=table}
\weightingablationcontent
\end{minipage}\vspace{12pt}}]
\else
\begin{table*}[t]
\weightingablationcontent
\end{table*}
\fi}
\weightingablationtable
\section{Ablation on the Reweighting Loss}
\label{app:reconstruction-ablation}

We compare four code:procedure:instruction weight ratios at
$\beta=0.25$: $1:1:1$, $1:2:3$, $1:2:5$, and $1:3:5$.
We additionally evaluate reconstruction without the auxiliary term
($\beta=0$) and $\beta=0.10$ with $1:2:3$ weights.

Table~\ref{tab:weighting-ablation} shows that the default setting
($\beta=0.25$, $1:2:3$) achieves the highest LLM-judge score,
$39.73/100$, exceeding the no-auxiliary-loss setting by 1.35 points
and $\beta=0.10$ by 0.68 points.
The $\beta=0.10$ setting has the highest Micro F1 and Skill-macro F1,
while omitting the auxiliary loss produces the shortest responses
and lowest average latency.
Among the $\beta=0.25$ weight configurations, uniform weighting
has the highest F1 scores and lowest latency.
We retain $\beta=0.25$ and $1:2:3$ as the default based on the
LLM-judge criterion; the tested higher-instruction-weight settings
do not improve this score.

\section{Skill-Content Retention in SHINE}
\label{app:shine-retention-analysis}

\paragraph{Diagnostic setup.}
We examine skill-content retention in
SHINE~\citep{liu2026shine} across checkpoints and
document-length cohorts using Qwen3-8B.

From the reconstruction records in SHINE's
skill-pretraining data, we sample 300 distinct skills:
150 from a broad-length baseline cohort and 150 from
a long-enriched cohort. Each cohort contains 50 skills
per content category.

Each skill contributes one contiguous code, procedure
(process), or instruction span, with target-span token
lengths approximately matched between cohorts within
each category.

For each skill, the complete document conditions
adapter generation. The frozen backbone then uses the
resulting adapter to reconstruct the document from a
reconstruction prompt under teacher forcing.

We first average the token-level negative log-likelihood
(NLL) within each target span and then average across
skills. All documents fit within the evaluation context
limits without truncation.

\begin{figure*}[t]
    \centering
    \includegraphics[
        width=0.96\textwidth
    ]{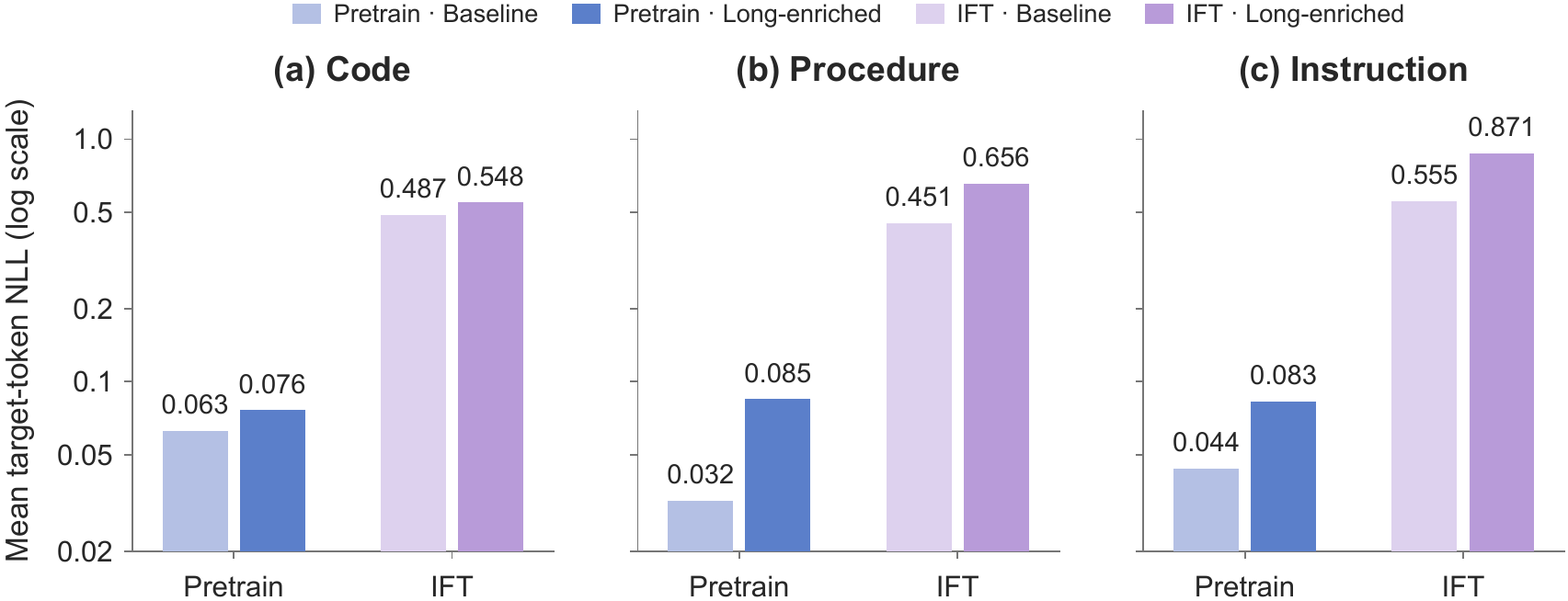}
    \caption{
        Skill-content reconstruction in SHINE using
        Qwen3-8B. Each bar reports the mean target-span
        token NLL across 50 skills, with adapters generated
        from the corresponding complete skill documents.
        The broad-length and long-enriched cohorts span
        117--1136 and 900--1139 tokens, respectively.
        The vertical axis uses a logarithmic scale;
        lower NLL indicates better conditional reconstruction.
    }
    \label{fig:shine-retention-length}
\end{figure*}

\paragraph{Checkpoint and cohort differences.}
Figure~\ref{fig:shine-retention-length} shows that the
IFT checkpoint has higher mean NLL than the pretraining
checkpoint in all six category--cohort combinations.
At both checkpoints, the long-enriched cohort also
has higher mean NLL in every content category.

For the IFT checkpoint, procedure NLL rises from
0.451 to 0.656 between the broad-length and
long-enriched cohorts (approximately 45\%), while
instruction NLL rises from 0.555 to 0.871
(approximately 57\%). The corresponding increase
for code is smaller, from 0.487 to 0.548
(approximately 13\%).

The pretraining checkpoint shows the same direction
of cohort differences, with mean NLL remaining
below 0.09 in all six groups.

\paragraph{Relation to our retention objective.}
These diagnostic results motivate explicitly
supervising the preservation of skill details,
particularly procedures and behavioral requirements
in longer documents.

Our structured retention objective
(Appendix~\ref{app:reconstruction-ablation})
assigns additional weight to code, procedure,
and instruction spans within full-document
reconstruction.

\section{Evaluation Details}
\label{app:evaluation-details}
\subsection{ALFWorld}
\label{app:alfworld-evaluation}

\subsection{Evaluation Protocol}
We evaluate Qwen3-8B, Qwen3.5-4B, and Qwen3.5-9B on all
274 ALFWorld episodes, comprising 140 \texttt{valid\_seen}
and 134 \texttt{valid\_unseen} episodes.
Within each rollout, every episode is evaluated once
using greedy decoding with temperature zero, seed 42, and a
limit of 50 environment steps.
Success is determined by the environment's \texttt{won} signal.
We pool the two splits to report success rates for each of the
six task types and overall, weighting each episode equally.

\subsection{SFT (LoRA) Baseline}
Starting from each original backbone, we train a rank-8 LoRA
adapter ($\alpha=8$) on 41,686 ALFWorld action-step examples
and use 1,661 examples for validation; these data exclude
evaluation games.
The backbone remains frozen, and only the current assistant's
reasoning and action are supervised, conditioned on the
interaction history.
All three models use AdamW with a learning rate of
$2\times10^{-5}$, a global batch size of 8, and seed 42
for one epoch.
We evaluate the final adapter.

\subsection{SHINE Baseline}
The SHINE checkpoints used in our experiments were originally
fine-tuned with thinking disabled.
To adapt them to our thinking-enabled ALFWorld evaluation,
we further train each checkpoint for one epoch on 2,000
complete samples drawn from our Stage~2 training data.
The backbone remains frozen; only the metanetwork,
Meta-LoRA, and memory tokens are updated.
Training uses a global batch size of 8, a learning rate of
$2\times10^{-5}$, and seed 42, yielding 250 optimizer updates.

\subsection{Skill Routing and Prompts}
No-Skill uses neither a skill document nor a skill adapter.
In-Context includes the task-specific skill document in the
first user message and retains it in subsequent histories.
SFT (LoRA) uses its shared trained adapter for all task types,
without including a textual skill document in the policy prompt.
SHINE and \textsc{ParametricSkills} supply skill information through
five LoRA adapters ($r=8$, $\alpha=8$), without including a
textual skill document in the policy prompt.
A fixed registry selects the adapter using the task type
provided by the ALFWorld game file; single-object and two-object
pick-and-place tasks share one adapter.

At each step, the policy receives the original task goal,
the current step index, the observation, and the complete set
of admissible actions.
The full interaction history, including prior reasoning and
actions, is retained without truncation.

\subsection{Reasoning and Action Handling}
The model is instructed to produce reasoning enclosed in
\texttt{<think>...</think>}, followed by one action enclosed in
\texttt{<action>...</action>}.
The SHINE runs, the Qwen3.5 baselines, and the Qwen3.5-4B
\textsc{ParametricSkills} run use the official thinking prefill
and extract a complete terminal action span outside reasoning.
The remaining Qwen3-8B runs and the Qwen3.5-9B
\textsc{ParametricSkills} run extract the last action span.
Action matching ignores case and redundant whitespace and maps
matching actions to the corresponding admissible commands.
Missing or empty actions default to \texttt{look}; nonempty
unmatched actions are passed directly to the environment,
without repair or resampling.

\subsection{SkillEval}

\label{app:skilleval-evaluation}

\paragraph{Tasks, repetitions, and metrics.}
Each reported method--backbone configuration is evaluated on the
same set of 100 SkillEval tasks.
We repeat the complete evaluation independently four times,
yielding 400 task-level runs per configuration.
We report the official overall SkillEval score and its K1/K2/K3
breakdowns as the mean $\pm$ standard deviation across the four
independent evaluation runs.

\paragraph{Agent environment and budgets.}
All methods use the same OpenCode agent harness, generation
limits, execution budget, and container resources.
Each agent execution runs in an isolated container.
Table~\ref{tab:skilleval-agent-config} lists the shared configuration.

\begin{table}[htbp]
\centering
\caption{Shared agent and container configuration for SkillEval.}
\label{tab:skilleval-agent-config}
\small
\setlength{\tabcolsep}{0pt}
\renewcommand{\arraystretch}{1.12}
\begin{tabular}{@{}>{\raggedright\arraybackslash}p{0.60\linewidth}
>{\raggedright\arraybackslash}p{0.40\linewidth}@{}}
\toprule
\textbf{Setting} & \textbf{Value} \\
\midrule
Agent harness & OpenCode v1.15.13 \\
Maximum context length & 40,960 tokens \\
Maximum output length & 8,192 tokens \\
Maximum agent steps per task & 20 \\
Wall-clock timeout per task & 800 seconds \\
CPU allocation per container & 2 cores \\
Memory allocation per container & 4 GiB \\
Maximum processes per container & 256 PIDs \\
\bottomrule
\end{tabular}
\end{table}

\paragraph{SFT (LoRA) baseline.}
We train a rank-8 LoRA adapter ($\alpha=16$) for each original
backbone on OpenCode trajectories, supervising only assistant
reasoning, responses, and tool calls.
All three models use 12,174 training and 713 validation examples.
Backbone weights remain frozen.
All three models use AdamW, a learning rate of $5\times10^{-5}$,
a global batch size of 8, and seed 42 for one epoch.
We evaluate the final adapters with textual skills available
to the OpenCode agent.
\section{Training Data Composition Across the Three Stages}
\label{app:three-stage-training-data}

Stage~1 uses reconstruction to support skill internalization;
Stages~2 and~3 both supervise skill use, without and with a specific
coding harness, respectively. Table~\ref{tab:three-stage-domain-composition}
summarizes the datasets' primary-use domains.


\subsection{Stage 1: Skill-Reconstruction Data}
\label{app:stage1-data-composition}
Each example conditions the hypernetwork on a skill document
and trains the adapted model to reconstruct that document.
The dataset contains 101{,}119 training and 5{,}316 validation
records, totaling 106{,}435 records.

\subsection{Stage 2: Skill-Exploitation Data without Harness}
\label{app:stage2-data-composition}
The Stage~2 dataset combines single-turn response examples and multi-turn
interaction sequences for skill exploitation. In both formats, skill
documents condition adapter generation separately from the supervised
dialogues.

\subsection{Stage 3: Skill-Exploitation Data with Harness}
\label{app:stage3-data-composition}
The Stage~3 dataset pairs skills with OpenCode agent
trajectories containing assistant reasoning, tool calls, environment
feedback, and final responses, providing supervision for skill use
within the agent harness.

\paragraph{Domain-labeling protocol.}
GLM-5.3-Flash assigns each unique \texttt{context} to one of 12 mutually
exclusive primary-use domains; records sharing a context inherit its
label. Table~\ref{tab:three-stage-domain-composition} pools all available
splits within each stage and counts repeated contexts as separate records.
Shares therefore describe records, not distinct skills.

\begin{table*}[t]
\centering
\caption{Primary-use domain distribution of the three training-stage
datasets. Counts pool all available splits; shares use each stage's
original-record total and may not sum to 100\% after rounding.}
\label{tab:three-stage-domain-composition}
\begingroup
\small
\setlength{\tabcolsep}{4pt}
\renewcommand{\arraystretch}{1.1}
\begin{tabular*}{\textwidth}{@{\extracolsep{\fill}}lrrrrrr@{}}
\toprule
& \multicolumn{2}{c}{Stage~1} & \multicolumn{2}{c}{Stage~2}
& \multicolumn{2}{c}{Stage~3} \\
\cmidrule(lr){2-3}\cmidrule(lr){4-5}\cmidrule(l){6-7}
Category & Records & Share (\%) & Records & Share (\%)
& Records & Share (\%) \\
\midrule
Other           & 1{,}778  & 1.67  & 44{,}574 & 38.73 & 51     & 0.40 \\
AI \& LLM       & 20{,}724 & 19.47 & 14{,}383 & 12.50 & 1{,}869 & 14.76 \\
API \& Auto     & 23{,}117 & 21.72 & 9{,}172  & 7.97  & 1{,}084 & 8.56 \\
Content \& Edu  & 7{,}319  & 6.88  & 3{,}820  & 3.32  & 470    & 3.71 \\
Product \& Biz  & 12{,}664 & 11.90 & 9{,}377  & 8.15  & 1{,}683 & 13.29 \\
Security        & 4{,}070  & 3.82  & 3{,}513  & 3.05  & 830    & 6.56 \\
Design \& Media & 5{,}357  & 5.03  & 3{,}090  & 2.69  & 303    & 2.39 \\
Software        & 14{,}957 & 14.05 & 12{,}788 & 11.11 & 3{,}073 & 24.28 \\
DevOps \& Cloud & 4{,}245  & 3.99  & 3{,}631  & 3.16  & 1{,}079 & 8.52 \\
Testing \& QA   & 2{,}582  & 2.43  & 2{,}563  & 2.23  & 843    & 6.66 \\
Web \& UI       & 5{,}142  & 4.83  & 4{,}974  & 4.32  & 480    & 3.79 \\
Data \& ML      & 4{,}480  & 4.21  & 3{,}194  & 2.78  & 894    & 7.06 \\
\midrule
Total           & 106{,}435 & 100.00 & 115{,}079 & 100.00
& 12{,}659 & 100.00 \\
\bottomrule
\end{tabular*}
\endgroup
\end{table*}

\newcommand{\examplefigure}[4][0.70\textheight]{%
    \begin{figure*}[p]
        \centering
        \includegraphics[
            width=0.96\textwidth,
            height=#1,
            keepaspectratio
        ]{#2}
        \caption{#3}
        \label{#4}
    \end{figure*}
}


\section{Templates and Examples}
\label{app:templates-examples}
\label{app:three-stage-examples}

This section provides representative examples of the
evaluation protocol and training data used in our framework.
We first present the LLM judge prompt and its scoring rubric,
followed by an example of a textual skill. We then illustrate
the three training stages through representative samples
covering skill reconstruction, single- and multi-turn skill
application, and agentic execution.


\begin{figure*}[p]
    \centering
    \includegraphics[width=0.96\textwidth,height=0.70\textheight,keepaspectratio]{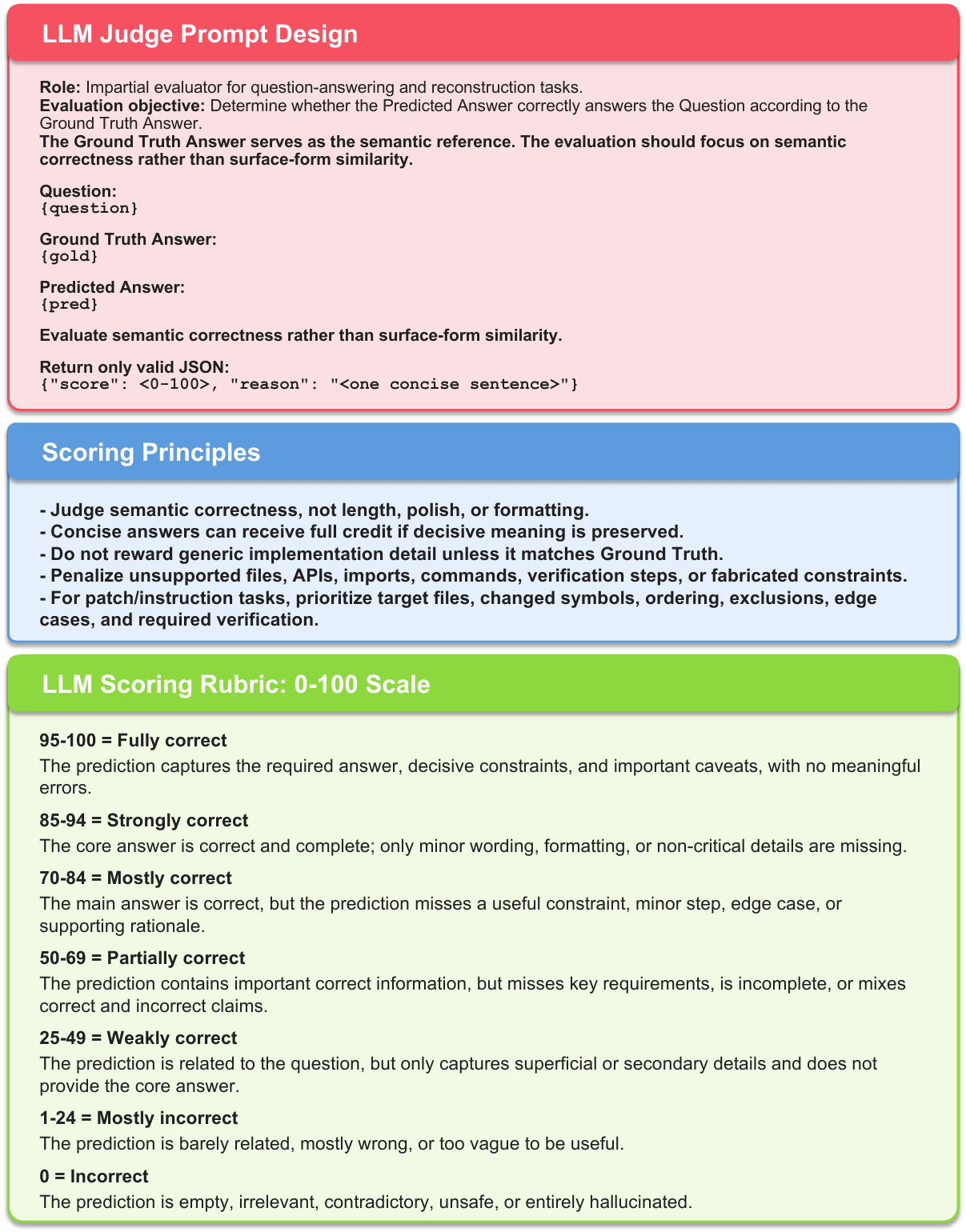}
    \caption{%
\textbf{LLM judge prompt and scoring rubric.}
The judge evaluates predicted responses against ground-truth
answers on a 0--100 scale, prioritizing semantic correctness
over surface-form similarity. The rubric penalizes unsupported
claims, fabricated constraints, omitted key requirements,
and incomplete or internally inconsistent responses.
}
    \label{fig:llm-judge-prompt}
\end{figure*}


\begin{figure*}[p]
    \centering
    \includegraphics[width=0.96\textwidth,height=0.70\textheight,keepaspectratio]{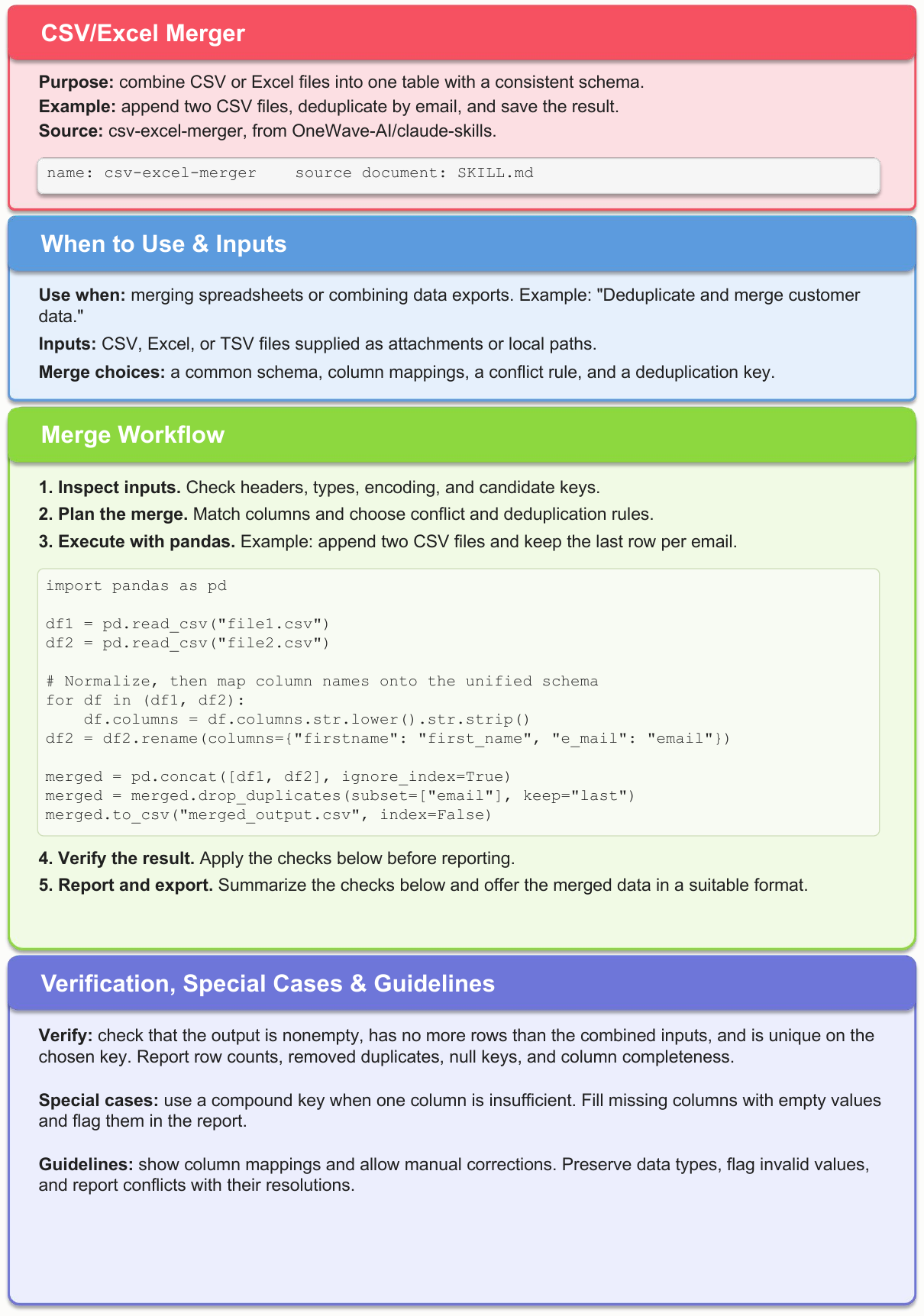}
    \caption{%
\textbf{Example of a textual skill for tabular data merging.}
The skill defines a reusable workflow covering input
requirements, schema normalization, field mapping,
deduplication, and output validation.
It combines two CSV files into a unified schema and
retains the last occurrence of each email address before
exporting the merged dataset.
}
    \label{fig:three-stage-skill-document}
\end{figure*}


\begin{figure*}[p]
    \centering
    \includegraphics[width=0.96\textwidth,height=0.70\textheight,keepaspectratio]{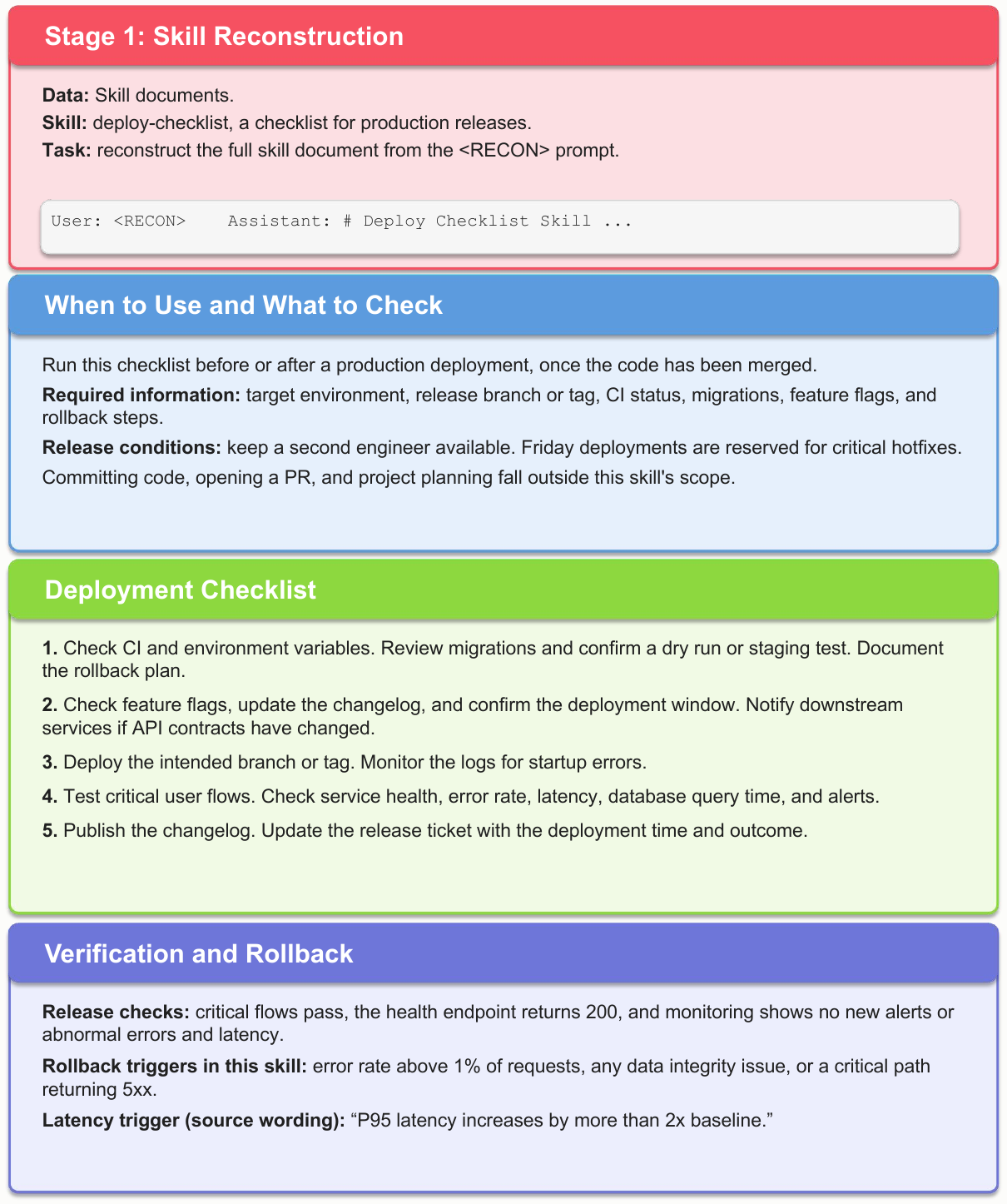}
    \caption{%
\textbf{Stage 1: Skill reconstruction.}
Given a \texttt{<RECON>} request, the model reconstructs
a complete skill document from the supervision target.
The example covers a deployment procedure, including
applicability conditions, verification checks, and rollback
criteria. It illustrates supervision over both procedural
knowledge and the conditions governing its use.
}
    \label{fig:three-stage-reconstruction}
\end{figure*}


\begin{figure*}[p]
    \centering
    \includegraphics[width=0.96\textwidth,height=0.68\textheight,keepaspectratio]{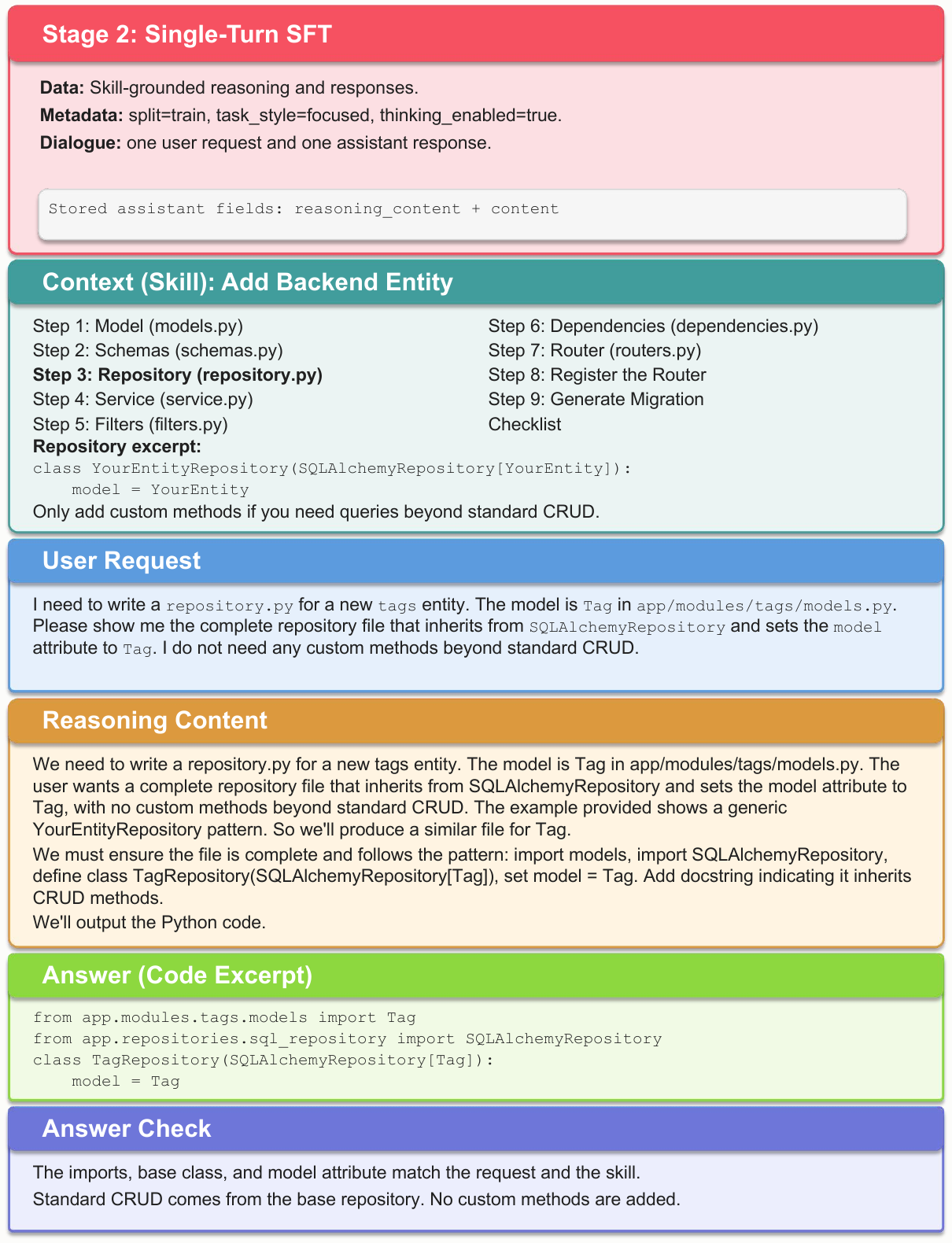}
    \caption{%
\textbf{Stage 2: Single-turn skill application.}
A user request is paired with an assistant response,
while the skill document is provided separately in the
record's \texttt{context} field.
Guided by repository-specific instructions, the assistant
implements the \texttt{Tag} entity, with its reasoning
and final code recorded separately.
}
    \label{fig:three-stage-single-turn}
\end{figure*}


\begin{figure*}[p]
    \centering
    \includegraphics[width=0.96\textwidth,height=0.68\textheight,keepaspectratio]{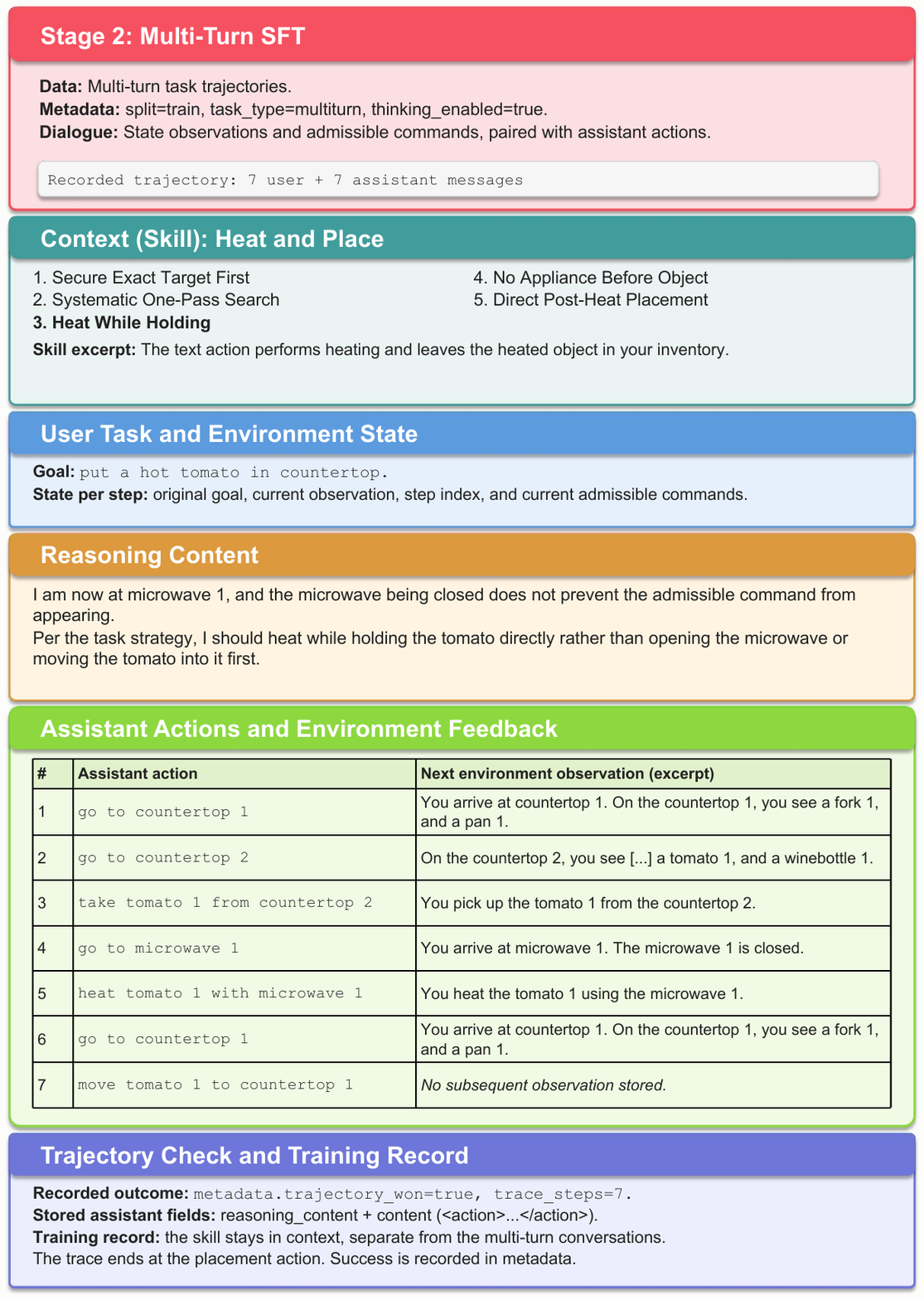}
    \caption{%
\textbf{Stage 2: Multi-turn skill application.}
A seven-step interaction trajectory collected using
the task environment's native execution harness.
Each user message specifies the task goal, current step,
observation, and admissible commands, while each assistant
message records its reasoning and selected action.
Guided by the heat-and-place skill, the assistant locates
a tomato, picks it up, heats it, and places it on a
countertop. Successful completion is recorded in
the trajectory metadata.
}
    \label{fig:three-stage-multi-turn}
\end{figure*}


\begin{figure*}[p]
    \centering
    \includegraphics[width=0.96\textwidth,height=0.70\textheight,keepaspectratio]{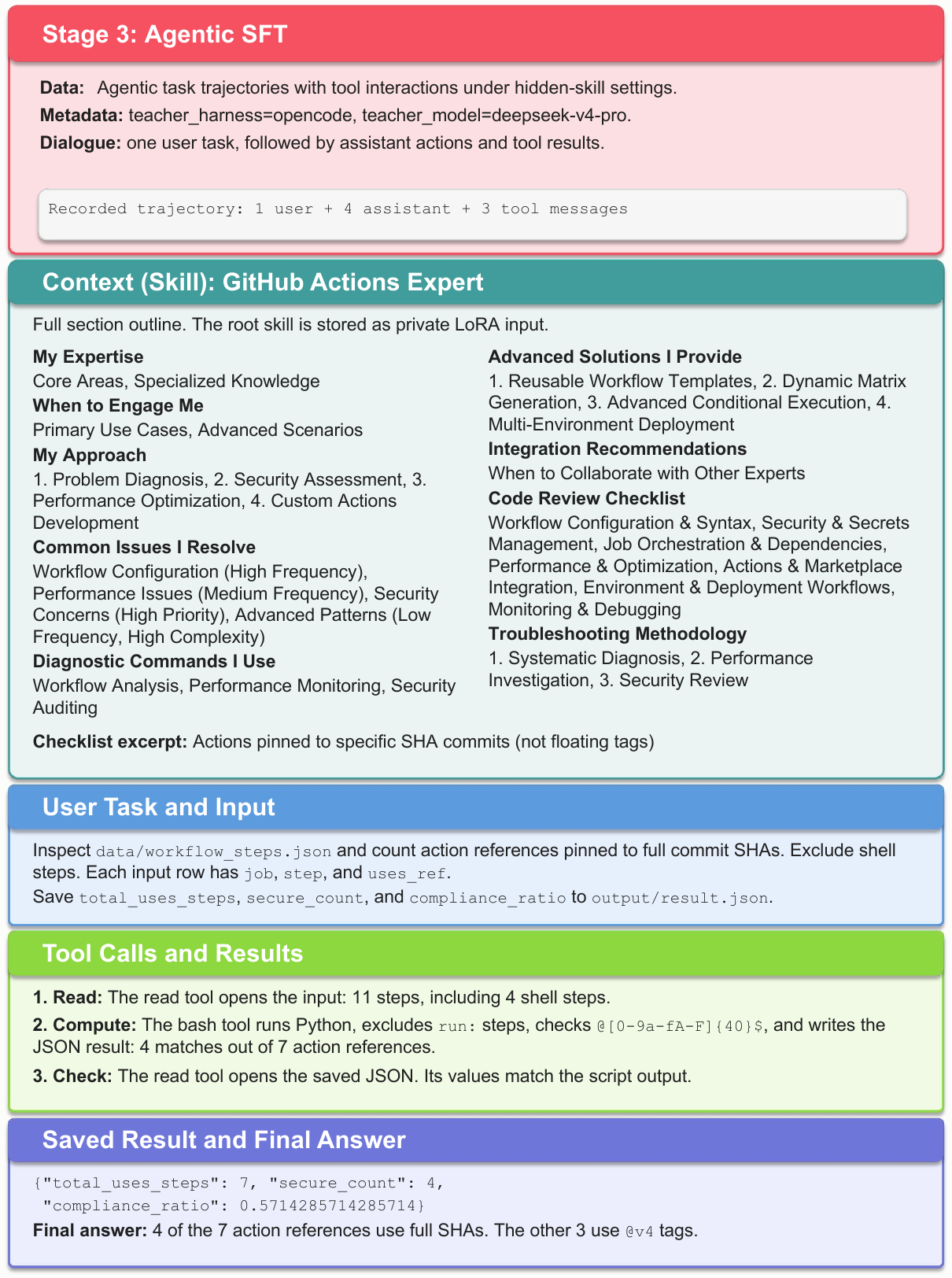}
    \caption{%
\textbf{Stage 3: Agentic execution.}
An agentic training trajectory for auditing GitHub Actions
references. The sample records successive assistant actions
and their corresponding tool outputs, preserving intermediate
observations alongside subsequent decisions.
It illustrates supervision over multi-step,
evidence-guided task execution rather than the final
response alone.
}
    \label{fig:three-stage-agentic}
\end{figure*}

\clearpage
\end{document}